\documentclass[10pt,twocolumn,letterpaper]{article}

 \usepackage[pagenumbers]{cvpr} % To force page numbers, e.g. for an arXiv version

\usepackage{amsmath}
\usepackage{amssymb}
\usepackage{graphicx}
\usepackage{array}
\usepackage{tabularx}

\usepackage{booktabs}   
\usepackage{multirow}   
\usepackage{makecell}   
\usepackage{adjustbox}  
\usepackage[table]{xcolor} 
\usepackage{colortbl} 
\usepackage{amsmath,amssymb}     
\usepackage[ruled]{algorithm2e}  

\definecolor{cvprblue}{rgb}{0.21,0.49,0.74}

\usepackage[pagebackref,breaklinks,colorlinks,allcolors=cvprblue]{hyperref}

\def\confName{CVPR}
\def\confYear{2027}

\title{Tell Models Where to Deviate: Fine-Grained Dynamic Residual Steering for Mitigating Hallucinations in Large Vision-Language Models}

\author{
  Jiahui Hu\textsuperscript{1},
  Xiaolong Luo\textsuperscript{2},
  Huamin Wang\textsuperscript{1},
  Fujun Han\textsuperscript{3}\thanks{Corresponding author.}\\
  \textsuperscript{1}Southwest University\quad
  \textsuperscript{2}City University of Hong Kong\\
  \textsuperscript{3}The Chinese University of Hong Kong, Shenzhen\\
  {\small\ttfamily hujiahui@email.swu.edu.cn, hanfujun@cuhk.edu.cn}
}
\begin{document}
\maketitle
\begin{abstract}
Object hallucination in large vision-language models (LVLMs) severely hinders their reliable deployment. Existing methods primarily rely on contrastive decoding and coarse-grained steering, yet they often lack fine-grained visual steering to adapt evolving visual attention across generation steps and representational differences across layers. To explore the in-depth causes of hallucinations and address this limitation, we propose a novel attention-dynamics-aware metric, Visual Attention Discrepancy (VAD), and reveal how visual deviation patterns vary across layers and heads. Besides, guided by these findings, we design an effective Dynamic Discrepancy-Aware Residual Steering framework, \textit{i.e.}, \textbf{DARES}, to achieve fine-grained visual steering in a training-free manner. Specifically, we first propose Discrepancy-Aware Adaptive Gating (DAAG), which combines shallow-layer visual attention discrepancy with generation history, to determine when to intervene. Then, we develop the Grounded Visual Deviation Steering (GVDS) to achieve token-level and layer-adaptive steering. Furthermore, a Dimension-wise Guidance Modulation (DGM) is introduced to coordinate the steering direction between the attention update and complete residual injection. Finally, extensive experiments and comparisons are conducted on five dominant LVLMs and four hallucination benchmarks. The results demonstrate that our DARES effectively reduces object hallucinations and exhibits promising performance across different models.
\end{abstract}   

\section{Introduction}
Large vision-language models (LVLMs) have achieved significant success in multi-modal tasks \cite{wang2024qwen2,chen2023shikra,liu2023visual,liu2024improved,zhu2024minigpt,dai2023instructblip,SOU,MLLMISU,OVID}, \textit{e.g.}, image captioning and visual question answering. %by integrating pretrained visual encoders with large language models. 
However, current LVLMs still have a basic and serious challenge, \textit{i.e.}, hallucination \cite{zechen2024hallucination,farquhar2024detecting,liu2024survey,min2025mitigating}. We can understand it as fluent but visually inconsistent descriptions. These errors undermine the credibility of generated content and limit the reliable use of these models in the real world. Therefore, mitigating hallucination is essential and necessary for the reliable deployment of LVLMs.

\begin{figure}[t]
    \centering
    \includegraphics[width=\columnwidth]{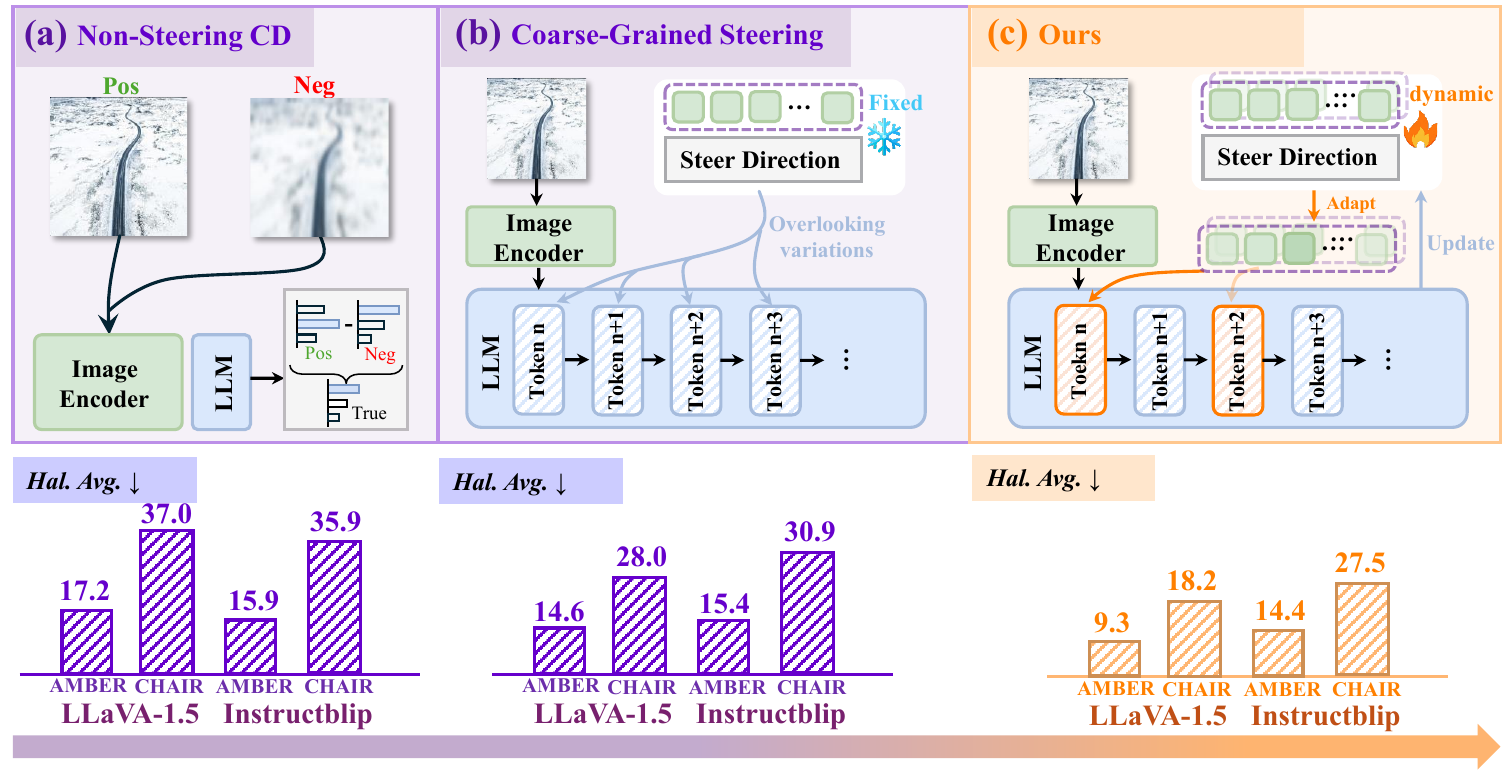}
    \caption{\textbf{Existing paradigms and our fine-grained dynamic steering.} (a) Non-steering contrastive decoding requires additional forward computation, incurring high inference overhead. (b) Coarse-grained steering reuses fixed directions that struggle to accommodate variations across generation steps and layers. (c) Our method enables selective residual steering that evolves with each token and adapts to each layer within a single forward pass, achieving better hallucination mitigation.}
    \label{fig:eff_time}
\par\vspace{-8pt}
\end{figure}
\begin{figure*}[t]
    \centering
    \includegraphics[width=\textwidth]{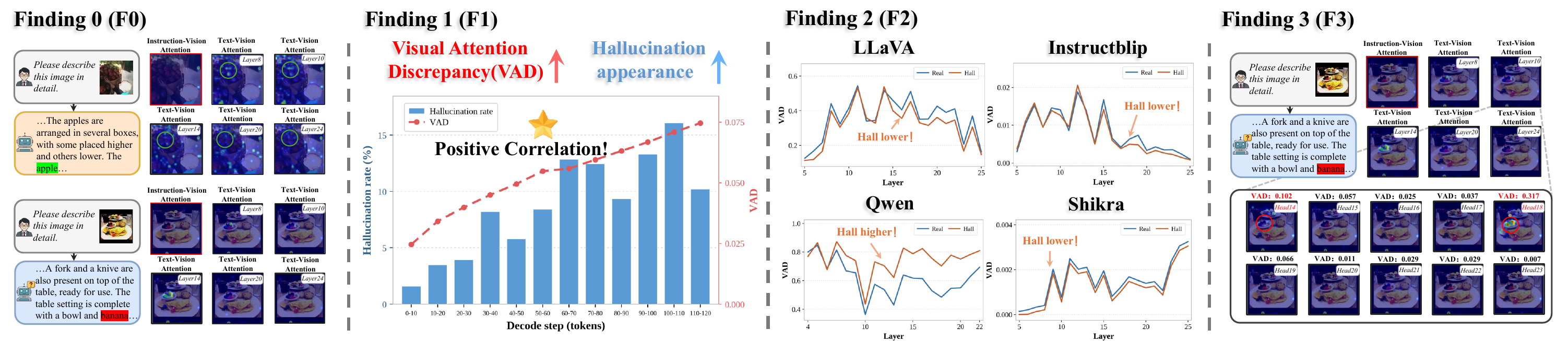}
    \caption{\textbf{Our four core findings.} (F0) Visual reference: TVA for grounded objects focuses on the target, whereas hallucinated objects largely retain IVA patterns, supporting IVA as a reference. (F1) Risk indication: Shallow-layer VAD increases during generation and positively correlates with hallucination rates. (F2) Layer and backbone dependence: The magnitude of VAD differences varies across layers, while their direction differs across backbones, supporting layer-adaptive steering. (F3) Head-level evidence: Some heads focus on objects, whereas others attend almost uniformly, suggesting that head-level differences encode visually grounded steering signals.}
    \label{fig:vad_analysis}
\end{figure*}

To mitigate hallucinations in LVLMs, current works have increasingly explored training-free approaches \cite{yin2024woodpecker,wei2024dopra,huang2024opera,an2025mitigating,zhang2026hallucination}. These works can be divided into non-steering and steering approaches. Non-steering methods typically intervene directly in logits or decoding strategies, with contrastive decoding (CD) comparing predictions from original and perturbed inputs~\cite{leng2024mitigating,wang2024mitigating,favero2024multi,huo2025self,gong2024damro,zhang2025self,chuang2024dola}. Although these works are effective, they require additional forward passes, which increases inference overhead. To solve these issues, some steering methods are proposed, \textit{e.g.}, MemVR \cite{zou2024look}, VTI \cite{liu2025reducing}, VISTA \cite{li2025hidden}, FLB \cite{ha2026first}, RUDDER \cite{zou2025adaptive}. Steering methods aim to inject steering vectors into the LLM within a single forward pass to guide generation, and they can effectively mitigate hallucinations caused by visual information decay. Despite these works achieving promising performance in mitigating hallucinations, they still raise basic and significant questions, \textit{i.e.}, existing methods primarily rely on coarse-grained steering with fixed vectors (i.e., applying uniform steering across all generation steps), overlooking the evolving visual reliance across decoding steps and the variations in visual representations across layers. As shown in Fig.~\ref{fig:eff_time}, we can find that contrastive decoding has low performance due to inadequate visual grounding, whereas coarse-grained steering reuses fixed directions with limited adaptability to evolving generation states. \par 

To address these questions, we first revisit hallucination through differences in visual attention distributions and define Visual Attention Discrepancy (VAD), a novel attention-dynamics-aware metric, to quantify and reveal how the Text--Vision Attention (TVA) distributions of generated tokens deviate from this reference across attention layers. As shown in Fig.~\ref{fig:vad_analysis}, we provide detailed analysis and the evolution process, \textit{i.e.}, Visual reference $\rightarrow$ Risk indication $\rightarrow$ Layer and backbone dependence $\rightarrow$ Head-level evidence. Through progressive analyses, we reveal a positive correlation between shallow-layer VAD and hallucination rates, along with backbone-dependent variations in VAD–factuality associations across layers. Furthermore, we find that distinct VAD patterns across attention heads suggest that different heads encode steering signals toward genuine visual evidence.

Motivated by these findings, to effectively mitigate hallucinations in LVLMs, we then propose Dynamic Discrepancy-Aware Residual Steering (DARES), a unified training-free framework for fine-grained dynamic steering. Specifically, we first propose Discrepancy-Aware Adaptive Gating (DAAG) to monitor hallucination risk and determine when to inject guidance, which can avoid disruptions to language fluency. Besides, to construct dynamic steering vectors, we design the Grounded Visual Deviation Steering (GVDS) method. GVDS can construct layer-specific steering directions that evolve with each token. Furthermore, we introduce Dimension-wise Guidance Modulation (DGM) to coordinate the integration of steering vectors with existing representations. Our DGM can not only achieve residual calibration but also can limit overcorrection of valid semantic representations. Finally, we conduct comprehensive experiments and comparisons to verify the effectiveness of the proposed framework and methods on multiple different benchmark datasets and architectures. The experimental results show that DARES can reach promising performance and effectively mitigate hallucinations in LVLMs. \par

In total, our contributions can be summarized as follows:
\begin{itemize}
    \item \textbf{Novel Diagnostic Perspective}. We rethink hallucinations from the perspective of visual attention discrepancies and introduce a novel diagnostic metric, \textit{i.e.}, Visual Attention Discrepancy (VAD), to systematically reveal their correlation with hallucination generation in depth.
    \item \textbf{Unified Mitigation Framework}. We propose DARES, a unified training-free framework with three effective designs, \textit{i.e.}, Discrepancy-Aware Adaptive Gating (DAAG), Grounded Visual Deviation Steering (GVDS), and Dimension-wise Guidance Modulation (DGM), to mitigate hallucinations in LVLMs.
    \item \textbf{Extensive Experiments and Validation}. Comprehensive experiments and comparisons are conducted across multiple dominant benchmarks and LVLM backbones. Experimental results demonstrate that our DARES significantly reduces object hallucinations and exhibits strong generalization across different architectures. 
\end{itemize}

\section{Related Work}

\noindent\textbf{Non-steering methods}. Non-steering methods typically intervene directly in logits or decoding strategies. Many improve visual grounding by recalibrating the final logits, but additional forward passes incur substantial latency. For example, VCD contrasts the prediction distributions of original and perturbed images \cite{leng2024mitigating}; PAI subtracts text-only logits while enhancing visual attention to mitigate text inertia \cite{liu2024paying}; ASCD combines internal attention intervention with contrastive decoding \cite{wang2026ascd}; and Self-Aug selects visual augmentations according to query semantics and adaptively adjusts candidate tokens using the output distribution \cite{im2026self}. Nevertheless, these methods often lack fine-grained visual steering, limiting hallucination mitigation performance, while additional forward passes increase inference overhead.

\noindent\textbf{Steering methods}. Steering methods inject steering vectors into the LLM during generation to guide its trajectory. Although they operate within a single forward pass and are more efficient than CD, earlier methods still incur high computational costs at runtime: ASD steers generation away from predefined hallucination directions \cite{su2025activation}, while VISTA injects signal vectors computed from activation differences \cite{li2025hidden}. More recent methods improve efficiency: FLB incorporates first-step logits into subsequent predictions \cite{ha2026first}, while RUDDER extracts a sample-specific visual direction from prefill residual updates with negligible additional overhead \cite{zou2025adaptive}. However, existing steering methods remain overly coarse-grained: they apply fixed steering at every generation step, overlooking evolving visual requirements and representational differences across layers, thereby limiting their effectiveness. Although RUDDER features relatively advanced designs, it mainly produces coarse-grained token-level scalar gates to control steering strength. In contrast, our method aggregates current visual information based on historical head-level discrepancies to construct steering directions that evolve with each token and adapt to each layer's representations. With selective gating and dimension-wise modulation, it performs fine-grained dynamic residual intervention in one forward pass.
\section{Motivation and Analysis}
\label{sec:analysis}

To examine differences in visual attention associated with different text segments, we distinguish Instruction-Vision Attention (IVA) from Text-Vision Attention (TVA) for generated text. We project the attention maps onto the image plane to inspect their layer-wise evolution for tokens referring to grounded and hallucinated objects. As shown in Fig.~\ref{fig:vad_analysis}~(F0), TVA for grounded objects consistently focuses on the target object in intermediate layers while retaining some response patterns shared with IVA. This indicates that the model preserves its initial visual focus while dynamically activating relevant local regions as the generation context evolves. In contrast, TVA for hallucinated objects closely resembles IVA across most layers, with additional local responses appearing only sporadically in a few layers. These observations indicate that IVA can serve as an initial, task-conditioned visual reference, while deviations of TVA from this reference provide useful cues for distinguishing grounded from hallucinated objects (F0).

To quantify these differences, we define Visual Attention Discrepancy (VAD) as the deviation of TVA from IVA:

\[
\mathrm{VAD}_{t}^{(\ell,h)}
=
1-
\left\langle
\mathcal N\!\left(\mathcal M\!\left(\mathbf A_{\mathrm{vis},t}^{(\ell,h)}\right)\right),
\mathcal N\!\left(\mathcal M\!\left(\mathbf A_{\mathrm{ins}}^{(\ell)}\right)\right)
\right\rangle .
\]

Here, \(\mathbf A_{\mathrm{vis},t}^{(\ell,h)}\in\mathbb R^{N_v}\) denotes the attention distribution over \(N_v\) visual tokens for the \(t\)-th generated token at head \(h\) of layer \(\ell\). The same-layer visual reference, \(\mathbf A_{\mathrm{ins}}^{(\ell)}\), is obtained during prefill by averaging attention over user instruction positions and attention heads. \(\mathcal M(\cdot)\) and \(\mathcal N(\cdot)\) denote per-map min-max normalization and \(L_2\) normalization, respectively. The inner product corresponds to the cosine similarity between the normalized attention maps. A higher VAD indicates a greater deviation of current visual attention from the initial reference.

\subsection{Sample-Level Correlation Analysis}

To investigate the correlation between VAD and hallucination, we randomly sample 500 images from MSCOCO \cite{lin2014microsoft} and measure hallucination rates and shallow-layer (L0--L2) VAD at different generation lengths. As shown in Fig.~\ref{fig:vad_analysis}~(F1), VAD rises sharply during generation, while hallucination frequency surges where visual consistency is weakest, indicating a positive correlation between the two. This reveals an accumulation of early visual mismatch that methods focusing solely on overall visual attention fail to capture. Nevertheless, further analysis of the 500 samples shows that shallow-layer VAD does not significantly affect the model's final prediction accuracy. These results suggest that, although shallow-layer VAD plays a role in model behavior and can help identify hallucinations, it is not the primary cause of hallucination.

\subsection{Layer-Wise Dynamic Analysis}

To further understand the visual discrepancy patterns of grounded and hallucinated objects, we quantitatively compare the VAD distributions of their associated tokens across backbones and intermediate layers using the same 500 images. Within a given backbone, VAD differences between the two token groups generally exhibit a relatively consistent direction (Fig.~\ref{fig:vad_analysis}~(F2)). However, the patterns differ markedly across backbones. In particular, the state-of-the-art Qwen2.5-VL exhibits a VAD pattern opposite to those of the other three backbones. This may be because the initial visual attention reflected in Qwen2.5-VL's IVA is already sufficiently focused, reducing the need for TVA to attend extensively to additional regions during generation. Moreover, the magnitude of VAD differences varies across layers. These results indicate that the association between VAD and object factuality depends on both the backbone and the layer (F2), providing a basis for constructing steering vectors tailored to each layer's representations.

\begin{figure*}[t]
    \centering
    \includegraphics[width=\textwidth]{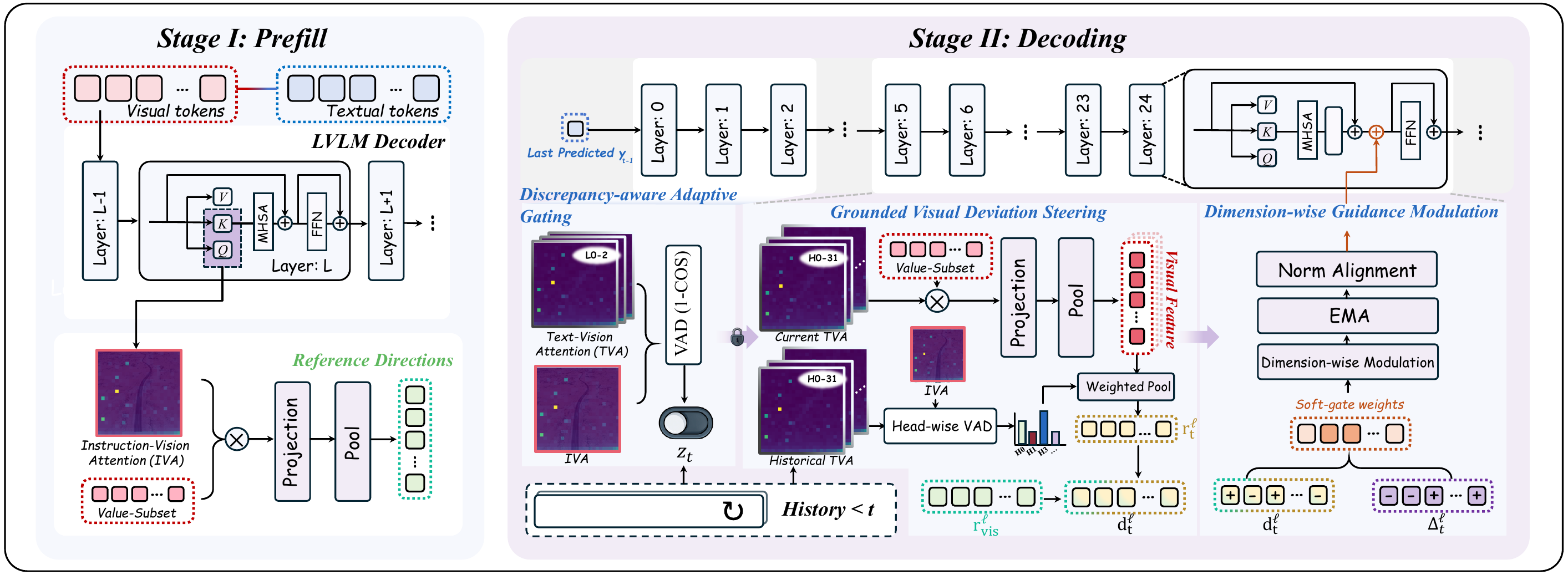}
   \caption{\textbf{Overview of DARES.} During prefill, we cache the instruction--vision reference and layer-specific reference directions. During decoding, DAAG combines shallow-layer VAD with generation history to determine when to intervene; GVDS aggregates current visual features according to historical head-level discrepancies and incorporates the instruction reference to construct token-wise updated, layer-adaptive steering directions; and DGM coordinates residual injection through dimension-wise modulation, smoothing, and norm alignment. The entire process requires only a single forward pass.}
    \label{fig:dares_framework}
\end{figure*}

\subsection{Head-Level Heterogeneity Analysis}

Since layer-level results reflect the aggregated behavior of multiple attention heads, we further examine head-wise VAD patterns and the corresponding spatial TVA distributions in LLaVA-1.5 to uncover the head-level differences underlying these overall trends. As shown in Fig.~\ref{fig:vad_analysis}~(F3), attention heads exhibit markedly different behaviors: some precisely attend to object-level regions, whereas others distribute attention almost uniformly across the image, exhibiting limited discriminative focus. Together with the layer-wise analysis (F2), this pronounced heterogeneity suggests that steering signals toward genuine visual evidence---namely, signals aligned with grounded visual deviation directions---are already intrinsically encoded within different attention heads during generation (F3).
\section{Methodology}
\label{sec:method}

\subsection{Preliminaries}
\label{sec:preliminaries}

Transformer-based LVLMs update representations through the residual stream \cite{vaswani2017attention}. We define the self-attention output after output projection as the attention residual update $\mathbf a_{\ell,t}$, representing the representation increment contributed by layer $\ell$ at decoding position $t$:

\begin{equation}
\mathbf a_{\ell,t}
=
\sum_{h=1}^{H}
\mathbf W_O^{(\ell,h)}
\left(
\sum_{j=1}^{N_t}
A_{t,j}^{(\ell,h)}\mathbf v_j^{(\ell,h)}
\right).
\label{eq:dares_attention_update}
\end{equation}
where $H$ is the number of heads, $N_t$ is the number of accessible tokens, and $A_{t,j}^{(\ell,h)}$ and $\mathbf v_j^{(\ell,h)}$ denote the attention weight and corresponding value vector, respectively.

We exploit this structure in two stages of autoregressive generation. During prefill, the model processes visual tokens and the text prompt, while DARES caches the instruction--vision attention reference and layer-specific instruction-conditioned reference vectors. During decoding, DARES monitors hallucination risk using shallow-layer VAD and selectively injects visual steering vectors into the hidden residual stream.

\subsection{DARES Overview}

To address these questions, in this paper we propose DARES, an effective training-free framework for fine-grained dynamic residual steering. As shown in Fig.~\ref{fig:dares_framework}, DARES resolves the three questions through DAAG (when to intervene), GVDS (how to construct vectors), and DGM (how to inject). Crucially, DARES reuses the model's visual representations and operates in a single forward pass. The detailed designs of the proposed methods are shown below.

\subsection{Discrepancy-Aware Adaptive Gating}
\label{sec:daag}

To determine when to intervene, we construct a history-adaptive gate based on shallow-layer VAD. Given the gating layer set $\mathcal G$ and injection layer set $\mathcal I$, we aggregate VAD across gating layers:

\begin{equation}
\overline{\mathrm{VAD}}_t
=
\frac{1}{|\mathcal G|}
\sum_{\ell\in\mathcal G}\mathrm{VAD}_t^{(\ell)}.
\label{eq:dares_aggregated_vad}
\end{equation}
where $\mathrm{VAD}_t^{(\ell)}$ is computed by comparing the head-averaged TVA with the same-layer IVA reference. To adapt to the current description's discrepancy level, we update the intervention threshold using generation history:

\begin{equation}
z_t
=
\mathbf{1}\left[
\overline{\mathrm{VAD}}_t>\lambda\mu_t
\right],
\label{eq:dares_gate}
\end{equation}
where $\mu_t=|\mathcal T_{<t}|^{-1}\sum_{k\in\mathcal T_{<t}}\overline{\mathrm{VAD}}_k$ is the mean historical aggregated VAD, $\mathcal T_{<t}$ denotes the set of generation positions preceding the current step, and $\lambda$ controls gating sensitivity. The gating decision is shared by subsequent injection layers, allowing discrepancy monitoring and residual intervention to proceed within the same forward pass.

\subsection{Grounded Visual Deviation Steering}
\label{sec:gvds}

Having determined when to intervene, we construct dynamic steering vectors by combining backbone-specific discrepancy patterns (F2) with head-level heterogeneity (F3), yielding directions that evolve with each token and adapt to each layer's representations.

\noindent\textbf{Attention-Weighted Visual Feature Aggregation.} We first average raw TVA and head-wise IVA across the gating layers, yielding $\bar{\mathbf A}_{\mathrm{vis},t}^{(h)}$ and $\bar{\mathbf A}_{\mathrm{ins}}^{(h)}$, respectively. For an injection layer $\ell\in\mathcal I$, we use these weights to aggregate its visual values and apply the output projection to obtain head-wise visual features:

\begin{equation}
\mathbf r_t^{(\ell,h)}
=
\mathbf W_O^{(\ell,h)}
\left(
\sum_{j\in\mathcal J_v}
\bar A_{\mathrm{vis},t,j}^{(h)}\mathbf v_j^{(\ell,h)}
\right).
\label{eq:dares_current_head_feature}
\end{equation}
where $\mathcal J_v$ is the set of visual token positions. The instruction-conditioned feature $\mathbf r_{\mathrm{ins}}^{(\ell,h)}$ is computed analogously and cached during prefill. Each layer uses its own values and output projection, yielding features adapted to its specific representations.

\noindent\textbf{Historical Discrepancy-Driven Head Fusion.} To exploit the head-level heterogeneity revealed by F3, we average $\bar{\mathbf A}_{\mathrm{vis},k}^{(h)}$ over historical positions $k\in\mathcal T_{<t}$ to obtain head-wise historical attention maps and compute their discrepancy $\mathrm{VAD}_{\mathrm{hist},t}^{(h)}$ relative to the shared IVA reference $\bar{\mathbf A}_{\mathrm{ins}}=H^{-1}\sum_h\bar{\mathbf A}_{\mathrm{ins}}^{(h)}$. To account for backbone dependence, we determine the polarity $\sigma_b\in\{-1,+1\}$ on an independent calibration set of 100 randomly sampled examples and fix it during inference. The head scores are defined as

\begin{equation}
q_t^{(h)}
=
\begin{cases}
\mathrm{VAD}_{\mathrm{hist},t}^{(h)}, & \sigma_b=-1,\\
\left[1-\mathrm{VAD}_{\mathrm{hist},t}^{(h)}\right]_+, & \sigma_b=+1.
\end{cases}
\label{eq:dares_head_scores}
\end{equation}
where $[x]_+=\max(x,0)$. With normalized weights $w_t^{(h)}=q_t^{(h)}/(\sum_{h'}q_t^{(h')}+\varepsilon)$, we fuse head-wise features:

\begin{equation}
\mathbf r_t^{(\ell)}
=
\mathcal N\!\left(
\sum_{h=1}^{H}
w_t^{(h)}\mathcal N\!\left(\mathbf r_t^{(\ell,h)}\right)
\right).
\label{eq:dares_fused_visual_direction}
\end{equation}
where $\mathcal N$ denotes $L_2$ normalization, and $\varepsilon$ ensures numerical stability. Historical attention determines the fusion weights, whereas current attention determines the visual content aggregated at the current step.

\noindent\textbf{Reference Direction Adjustment.} Inspired by CD, we aggregate instruction-conditioned features into a fixed reference direction to adjust the current visual direction:

\begin{equation}
\mathbf d_t^{(\ell)}
=
\mathcal N\!\left(
\mathbf r_t^{(\ell)}
+\sigma_b\alpha\,\mathbf r_{\mathrm{ins}}^{(\ell)}
\right).
\label{eq:dares_candidate_direction}
\end{equation}
where $\mathbf r_{\mathrm{ins}}^{(\ell)}=\mathcal N\!\left(H^{-1}\sum_{h=1}^{H}\mathbf r_{\mathrm{ins}}^{(\ell,h)}\right)$, and $\alpha$ controls the reference contribution. The reference is fixed after prefill; the current visual direction evolves with generation.

\subsection{Dimension-wise Guidance Modulation}
\label{sec:dgm}

To control how guidance is injected, we coordinate the steering direction with the current attention update for residual injection.

\noindent\textbf{Dimension-wise Agreement Modulation.} We compute dimension-wise modulation weights using the element-wise product of the normalized steering direction and the normalized current attention update:

\begin{equation}
\mathbf g_t^{(\ell)}
=
\frac12\left[
\mathbf 1+
\tanh\!\left(
d_{\mathrm{model}}\,\mathbf d_t^{(\ell)}
\odot\mathcal N(\mathbf a_{\ell,t})
\right)
\right].
\label{eq:dares_dimension_gate}
\end{equation}
where $d_{\mathrm{model}}$ is the hidden dimension, and $\odot$ denotes element-wise multiplication. Same-sign components receive higher weights, while opposite-sign components are suppressed, aligning visual correction with current semantics and mitigating perturbations to valid representations.

\noindent\textbf{Temporal Smoothing and Residual Injection.} To mitigate directional fluctuations between successive steps, we maintain an independent exponential moving average (EMA) state for each injection layer and align the norm of the smoothed direction with that of the current attention update before injection:

\begin{equation}
\widetilde{\mathbf d}_t^{(\ell)}
=
\beta\!\left(
\mathbf g_t^{(\ell)}\odot\mathbf d_t^{(\ell)}
\right)
+
(1-\beta)\widetilde{\mathbf d}_{t-1}^{(\ell)}.
\label{eq:dares_ema}
\end{equation}

\begin{equation}
\mathbf a'_{\ell,t}
=
\mathbf a_{\ell,t}
+
z_t s\|\mathbf a_{\ell,t}\|_2\,
\mathcal N\!\left(\widetilde{\mathbf d}_t^{(\ell)}\right).
\label{eq:dares_residual_injection}
\end{equation}
where $\beta$ weights the current modulated vector, and $s$ controls steering strength. Injection layers share the gating signal but use their own steering directions and smoothing states, modifying only the attention residual update at the current decoding position.
\section{Experiments}
\subsection{Experimental Setup}

\noindent\textbf{Implementation Details.} We evaluate the effectiveness and generalizability of DARES on five mainstream LVLMs: LLaVA-1.5, InstructBLIP, Qwen2.5-VL, Shikra, and LLaVA-NeXT, all using their 7B versions. The experiments cover four standard hallucination benchmarks: AMBER for comprehensive multi-scenario evaluation, CHAIR for object hallucination in image captioning, OPOPE for object existence verification, and Object HalBench for object hallucination assessment via GPT-assisted object extraction. All models use greedy decoding with a maximum generation length of 512 tokens. We apply steering to layers 5--24 across all models. By default, the reference direction coefficient $\alpha$, steering strength \(s\), gating threshold coefficient \(\lambda\), and EMA update coefficient \(\beta\) are set to \(0.1\), \(2.5\), \(0.8\), and \(0.3\), respectively. The only backbone-specific adjustments are: \(\lambda=0.7\) for LLaVA-1.5, \(\alpha=0.8\) for LLaVA-NeXT, and \(\alpha=0.5\) and \(s=2.0\) for Qwen2.5-VL.

\noindent\textbf{Comparison Methods and Baselines.} We compare DARES with 10 representative training-free hallucination mitigation methods. These include steering-based methods (MemVR, VISTA, FLB, and RUDDER), as well as non-steering methods encompassing contrastive decoding and attention manipulation (VCD, ONLY \cite{wan2025only}, ASCD, Self-Aug, VisAttnSink \cite{kang2025see}, and AdaIAT \cite{zhong2026adaiat}).

\subsection{Main Results on AMBER and CHAIR}

Table~\ref{tab:main_results} presents the main results of DARES on AMBER \cite{wang2023amber} and the MSCOCO-based CHAIR benchmark \cite{rohrbach2018object}. To comprehensively compare hallucination performance, we additionally report the Hallucination Average (Hal.~Avg.), computed as the arithmetic mean of CHAIR, Hal, and Cog on AMBER, and of \(\mathrm{CHAIR}_s\) and \(\mathrm{CHAIR}_i\) on CHAIR. Lower values indicate a lower overall hallucination level. DARES achieves the lowest Hal.~Avg. across all five backbones on both benchmarks, outperforming the static steering and non-steering methods in the table. More importantly, the AMBER results show that DARES maintains over 85\% of the object coverage (Cover) of each original backbone while reducing hallucinations, with Cover even improving on InstructBLIP. This demonstrates a favorable trade-off between hallucination suppression and information coverage. The CHAIR results further confirm consistent reductions in both sentence-level and object-level hallucination rates across all five backbones. These consistent improvements across architectures and benchmarks collectively support the effectiveness and generalizability of DARES in mitigating object hallucinations in LVLMs.

\begin{table}[t]
\centering
\caption{\textbf{Evaluation of five LVLMs on AMBER and CHAIR (unit: \%).} Hal. Avg. is the arithmetic mean of each benchmark's hallucination metrics. For each metric and backbone, the best results are bold and the second-best underlined.}
\label{tab:main_results}
\setlength{\tabcolsep}{1.5pt} 
\scriptsize 
\resizebox{\columnwidth}{!}{ 
\begin{tabular}{@{}llcccccccc@{}}
\toprule

\multirow{2}{*}[-0.5ex]{Model} & \multirow{2}{*}[-0.5ex]{Method} & \multicolumn{5}{c}{AMBER} & \multicolumn{3}{c}{CHAIR} \\
\cmidrule(lr){3-7} \cmidrule(lr){8-10}
& & CHAIR\(\downarrow\) & Cover\(\uparrow\) & Hal\(\downarrow\) & Cog\(\downarrow\) & \makecell{Hal.\\Avg.\(\downarrow\)} & \(\mathrm{CHAIR}_s\)\(\downarrow\) & \(\mathrm{CHAIR}_i\)\(\downarrow\) & \makecell{Hal.\\Avg.\(\downarrow\)} \\
\midrule
\multirow{12}{*}{LLaVA-1.5} 
 & Vanilla & 7.3 & 50.9 & 33.2 & 3.9 & 14.8 & 52.2 & 14.5 & 33.4 \\
 & VCD & 8.3 & \textbf{53.1} & 39.7 & 3.6 & 17.2 & 58.2 & 15.7 & 37.0 \\
 & MemVR & 7.9 & 52.1 & 37.4 & 4.3 & 16.5 & 46.6 & 13.0 & 29.8 \\
 & ONLY & 10.4 & 48.7 & 42.4 & 4.5 & 19.1 & 49.8 & 14.3 & 32.1 \\
 & VISTA & - & - & - & - & - & 38.6 & 11.4 & 25.0 \\
 & ASCD & 6.0 & 51.1 & 30.2 & 2.7 & 13.0 & 44.0 & 12.3 & 28.2 \\
 & Self-Aug & 9.5 & 51.3 & 41.8 & 4.4 & 18.6 & 55.8 & 15.8 & 35.8 \\
 & VisAttnSink & 8.2 & \underline{52.8} & 38.6 & 3.9 & 16.9 & 43.2 & 13.8 & 28.5 \\
 & AdaIAT & 5.4 & 47.6 & \underline{23.2} & 2.4 & \underline{10.3} & \underline{36.6} & \underline{10.3} & \underline{23.5} \\
 & FLB & \underline{4.9} & 48.8 & 25.2 & \underline{2.3} & 10.8 & 43.5 & 12.0 & 27.8 \\
 & RUDDER & 7.4 & 50.1 & 32.9 & 3.7 & 14.6 & 42.4 & 13.6 & 28.0 \\
\rowcolor{blue!5}
 & DARES & \textbf{2.9} & 50.3 & \textbf{22.9} & \textbf{2.1} & \textbf{9.3} & \textbf{27.8} & \textbf{8.6} & \textbf{18.2} \\
\midrule
\multirow{7}{*}{InstructBLIP} 
 & Vanilla & 18.5 & 53.5 & 57.7 & 6.5 & 27.6 & 56.2 & 24.8 & 40.5 \\
 & VCD & 9.5 & 53.8 & 44.4 & 4.6 & 19.5 & 54.0 & 17.8 & 35.9 \\
 & ONLY & 14.4 & \underline{54.6} & 49.3 & 7.0 & 23.6 & 59.8 & 21.2 & 40.5 \\
 & Self-Aug & 9.3 & 54.0 & 44.9 & 4.6 & 19.6 & 59.4 & 17.1 & 38.3 \\
 & FLB & 9.0 & 53.6 & 43.8 & 4.7 & 19.2 & 52.5 & 15.8 & 34.2 \\
 & RUDDER & \underline{7.4} & 53.7 & \underline{35.0} & \underline{3.7} & \underline{15.4} & \underline{47.6} & \underline{14.1} & \underline{30.9} \\
\rowcolor{blue!5}
 & DARES & \textbf{5.8} & \textbf{55.2} & \textbf{33.9} & \textbf{3.6} & \textbf{14.4} & \textbf{43.4} & \textbf{11.6} & \textbf{27.5} \\
\midrule
\multirow{6}{*}{Qwen2.5-VL} 
 & Vanilla & 5.0 & \textbf{64.8} & 27.0 & 1.8 & 11.3 & 36.2 & 9.6 & 22.9 \\
 & MemVR & \underline{4.8} & \underline{64.3} & \underline{25.7} & 1.5 & \underline{10.7} & 33.8 & 9.6 & 21.7 \\
 & ASCD & 7.8 & 50.5 & 27.6 & \textbf{0.9} & 12.1 & \textbf{26.4} & 9.5 & 18.0 \\
 & AdaIAT & - & - & - & - & - & 28.4 & \underline{6.9} & \underline{17.7} \\
 & RUDDER & 5.0 & \textbf{64.8} & 28.0 & 1.6 & 11.5 & 37.0 & 9.6 & 23.3 \\
\rowcolor{blue!5}
 & DARES & \textbf{3.8} & 56.4 & \textbf{20.5} & \underline{1.0} & \textbf{8.4} & \underline{26.6} & \textbf{6.7} & \textbf{16.7} \\
\midrule
\multirow{6}{*}{Shikra} 
 & Vanilla & 9.7 & 51.2 & 45.5 & 5.0 & 20.1 & \underline{49.6} & 14.0 & \underline{31.8} \\
 & VCD & 9.7 & 51.2 & 45.5 & 5.0 & 20.1 & \underline{49.6} & 14.0 & \underline{31.8} \\
 & ONLY & 10.1 & 49.4 & 47.6 & 5.0 & 20.9 & 54.0 & 15.3 & 34.7 \\
 & ASCD & \underline{9.1} & \textbf{53.1} & 44.6 & \textbf{4.1} & \underline{19.3} & 50.2 & \underline{13.7} & 32.0 \\
 & Self-Aug & 9.6 & \underline{51.5} & \underline{44.3} & \underline{4.2} & 19.4 & 51.4 & \underline{13.7} & 32.6 \\
\rowcolor{blue!5}
 & DARES & \textbf{8.1} & 50.0 & \textbf{40.3} & \textbf{4.1} & \textbf{17.5} & \textbf{43.4} & \textbf{12.5} & \textbf{28.0} \\
\midrule
\multirow{4}{*}{LLaVA-NeXT} 
 & Vanilla & 8.7 & \underline{62.3} & 48.6 & 4.5 & 20.6 & \underline{27.0} & 7.6 & \underline{17.3} \\
 & VCD & 8.3 & \textbf{62.4} & 48.9 & \textbf{3.1} & 20.1 & 39.4 & 9.6 & 24.5 \\
 & MemVR & \underline{8.1} & 61.9 & \underline{46.8} & 4.4 & \underline{19.8} & 28.8 & \textbf{7.1} & 18.0 \\
\rowcolor{blue!5}
 & DARES & \textbf{5.5} & 61.7 & \textbf{39.6} & \underline{3.4} & \textbf{16.2} & \textbf{25.8} & \underline{7.2} & \textbf{16.5} \\
\bottomrule
\end{tabular}
}
\end{table}

\subsection{Generalization Experiments}

To assess generalizability beyond AMBER and CHAIR, we evaluate two complementary hallucination benchmarks: OPOPE \cite{chen2024halc} and Object HalBench \cite{rohrbach2018object}, which assess object existence verification and hallucination via GPT-assisted object extraction \cite{yu2024rlhf}, respectively.

\noindent\textbf{OPOPE.} As shown in Table~\ref{tab:generalization}, DARES improves the mean precision and \(F_{\beta=0.2}\) across all five backbones, with accuracy also improving on most backbones. These results indicate that its effectiveness extends from open-ended description generation to object existence verification, supporting its use for hallucination mitigation across task formats.

\noindent\textbf{Object HalBench.} As shown in Table~\ref{tab:generalization}, DARES reduces the sentence-level hallucination rate (C-SR) by 19.4\%, 15.6\%, 8.3\%, 49.6\%, and 2.8\% compared to the original LLaVA-1.5, InstructBLIP, LLaVA-NeXT, Qwen2.5-VL, and Shikra models, respectively, while also reducing C-s and C-i across all five backbones. Consistent improvements in all three hallucination metrics further support the effectiveness and generalizability of DARES across different model architectures and evaluation protocols.

\begin{table}[t]
\centering
\par\vspace{-4pt}
\caption{\textbf{Evaluation of five LVLMs on OPOPE and Object HalBench.} OPOPE reports accuracy, precision, and $F_{\beta=0.2}$; Object HalBench reports object hallucination and coverage metrics. All values except Num are percentages. The best results for each metric and backbone are bold.}
\label{tab:generalization}
\fontsize{6}{7}\selectfont 
\renewcommand{\arraystretch}{0.85} 
\setlength{\tabcolsep}{1pt} 
\begin{tabularx}{\columnwidth}{@{}ll*{7}{>{\centering\arraybackslash}X}@{}}
\toprule
\multirow{2}{*}[-0.5ex]{Model} & \multirow{2}{*}[-0.5ex]{Method} & \multicolumn{3}{c}{OPOPE} & \multicolumn{4}{c}{Object HalBench} \\
\cmidrule(lr){3-5} \cmidrule(lr){6-9}
& & Acc\(\uparrow\) & Pre\(\uparrow\) & \(F_{\beta=0.2}\uparrow\) & C-s\(\downarrow\) & C-SR\(\downarrow\) & C-i\(\downarrow\) & Num\(\uparrow\) \\
\midrule
\multirow{2}{*}{LLaVA-1.5} 
 & Vanilla & 84.7 & 91.6 & 91.2 & 47.7 & 49.5 & 26.2 & \textbf{289} \\
 & \cellcolor{blue!5}DARES & \cellcolor{blue!5}\textbf{84.9} & \cellcolor{blue!5}\textbf{92.3} & \cellcolor{blue!5}\textbf{91.5} & \cellcolor{blue!5}\textbf{38.0} & \cellcolor{blue!5}\textbf{39.9} & \cellcolor{blue!5}\textbf{20.3} & \cellcolor{blue!5}286 \\
\midrule
\multirow{2}{*}{InstructBLIP} 
 & Vanilla & 80.1 & 90.5 & 89.3 & 21.7 & 23.7 & 15.9 & 274 \\
 & \cellcolor{blue!5}DARES & \cellcolor{blue!5}\textbf{82.0} & \cellcolor{blue!5}\textbf{91.7} & \cellcolor{blue!5}\textbf{90.6} & \cellcolor{blue!5}\textbf{19.0} & \cellcolor{blue!5}\textbf{20.0} & \cellcolor{blue!5}\textbf{13.0} & \cellcolor{blue!5}\textbf{285} \\

\midrule
\multirow{2}{*}{Qwen2.5-VL} 
 & Vanilla & \textbf{81.9} & 94.7 & 93.2 & 12.3 & 13.3 & 6.8 & \textbf{278} \\
 & \cellcolor{blue!5}DARES & \cellcolor{blue!5}78.4 & \cellcolor{blue!5}\textbf{95.8} & \cellcolor{blue!5}\textbf{93.6} & \cellcolor{blue!5}\textbf{5.3} & \cellcolor{blue!5}\textbf{6.7} & \cellcolor{blue!5}\textbf{4.3} & \cellcolor{blue!5}238 \\
\midrule
\multirow{2}{*}{Shikra} 
 & Vanilla & 82.4 & 91.7 & 90.7 & 22.0 & 24.6 & 15.7 & \textbf{268} \\
 & \cellcolor{blue!5}DARES & \cellcolor{blue!5}\textbf{82.8} & \cellcolor{blue!5}\textbf{93.6} & \cellcolor{blue!5}\textbf{92.5} & \cellcolor{blue!5}\textbf{21.3} & \cellcolor{blue!5}\textbf{23.9} & \cellcolor{blue!5}\textbf{14.4} & \cellcolor{blue!5}\textbf{268} \\
\midrule
\multirow{2}{*}{LLaVA-NeXT} 
 & Vanilla & \textbf{79.5} & 94.6 & 92.8 & 13.7 & 14.5 & 8.0 & \textbf{282} \\
 & \cellcolor{blue!5}DARES & \cellcolor{blue!5}79.3 & \cellcolor{blue!5}\textbf{95.1} & \cellcolor{blue!5}\textbf{93.2} & \cellcolor{blue!5}\textbf{12.3} & \cellcolor{blue!5}\textbf{13.3} & \cellcolor{blue!5}\textbf{6.9} & \cellcolor{blue!5}279 \\
\bottomrule
\end{tabularx}

\end{table}

\subsection{Ablation Study}

\begin{figure}[t]
    \centering
    \includegraphics[width=\columnwidth]{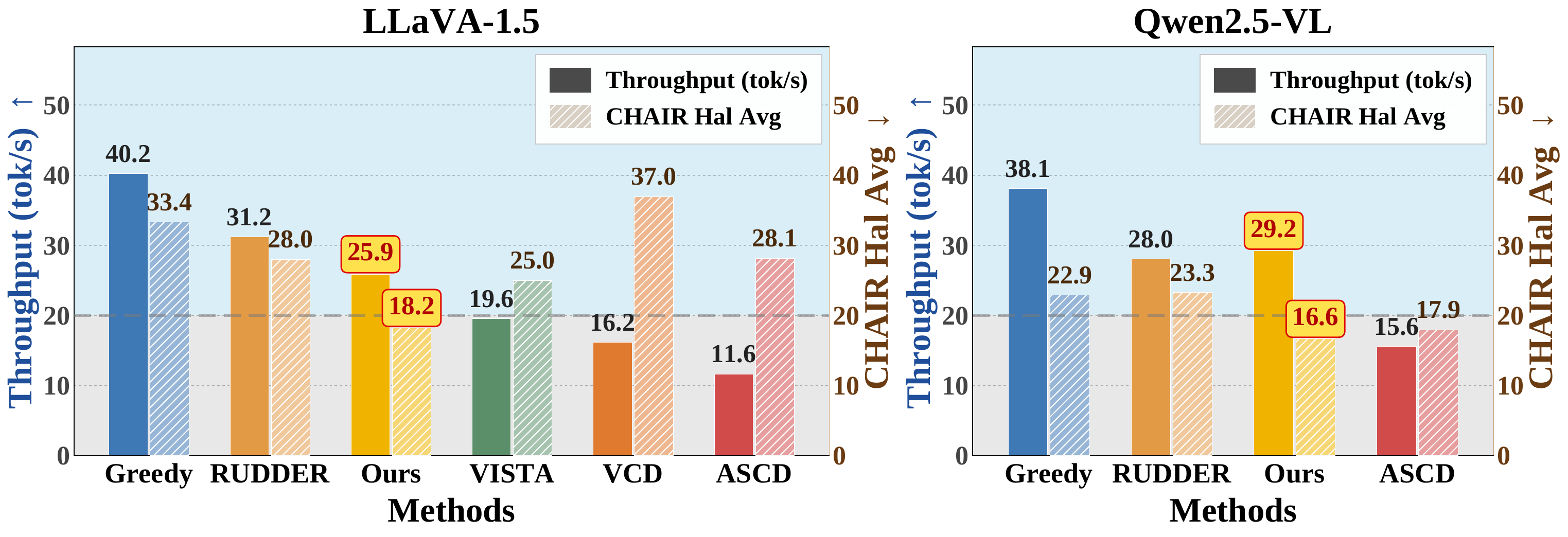}
    \caption{\textbf{Inference efficiency and hallucination comparison across decoding strategies on CHAIR.} Throughput measures inference speed; Hal. Avg. measures hallucination. On LLaVA-1.5, DARES achieves approximately \(1.6\times\) the throughput of VCD with fewer hallucinations. On Qwen2.5-VL, DARES improves throughput while reducing Hal.~Avg. compared with RUDDER.}
    \label{fig:efficiency}
    \par\vspace{-8pt}
\end{figure}

Table~\ref{tab:ablation} presents the ablation results of the three core components of DARES. We assess their contributions by progressively enabling Grounded Visual Deviation Steering (GVDS), Dimension-wise Guidance Modulation (DGM), and Discrepancy-Aware Adaptive Gating (DAAG). Enabling GVDS alone substantially reduces hallucination but also yields a considerable decline in object coverage, indicating that steering vector injection alone still incurs information coverage loss. Adding DGM mitigates this loss and further reduces hallucination, suggesting that dimension-wise modulation helps coordinate the steering direction with the current attention update, thereby reducing unnecessary representational perturbations. Further incorporating DAAG achieves the best overall balance between hallucination suppression and ground-truth object coverage. These results confirm the complementary roles of the three components and validate the rationale behind our unified design.

\subsection{Efficiency Analysis}

As shown in Fig.~\ref{fig:efficiency}, we compare throughput and Hal.~Avg. on the CHAIR generation task to assess the trade-off between inference efficiency and hallucination suppression. Compared with RUDDER, which uses a fixed source of guidance, DARES provides token-level, layer-adaptive dynamic steering. On LLaVA-1.5, it reduces Hal.~Avg. by 35.0\% with only an approximately 17.0\% decrease in throughput; on Qwen2.5-VL, it improves both throughput and hallucination suppression. Compared with VCD, which requires additional forward passes, DARES achieves approximately $1.6\times$ its throughput on LLaVA-1.5 while attaining a lower Hal.~Avg. These results show that fine-grained dynamic steering in one forward pass balances inference efficiency and hallucination mitigation.

\subsection{Discussion}

\begin{figure}[t]
    \centering
    \includegraphics[width=\columnwidth]{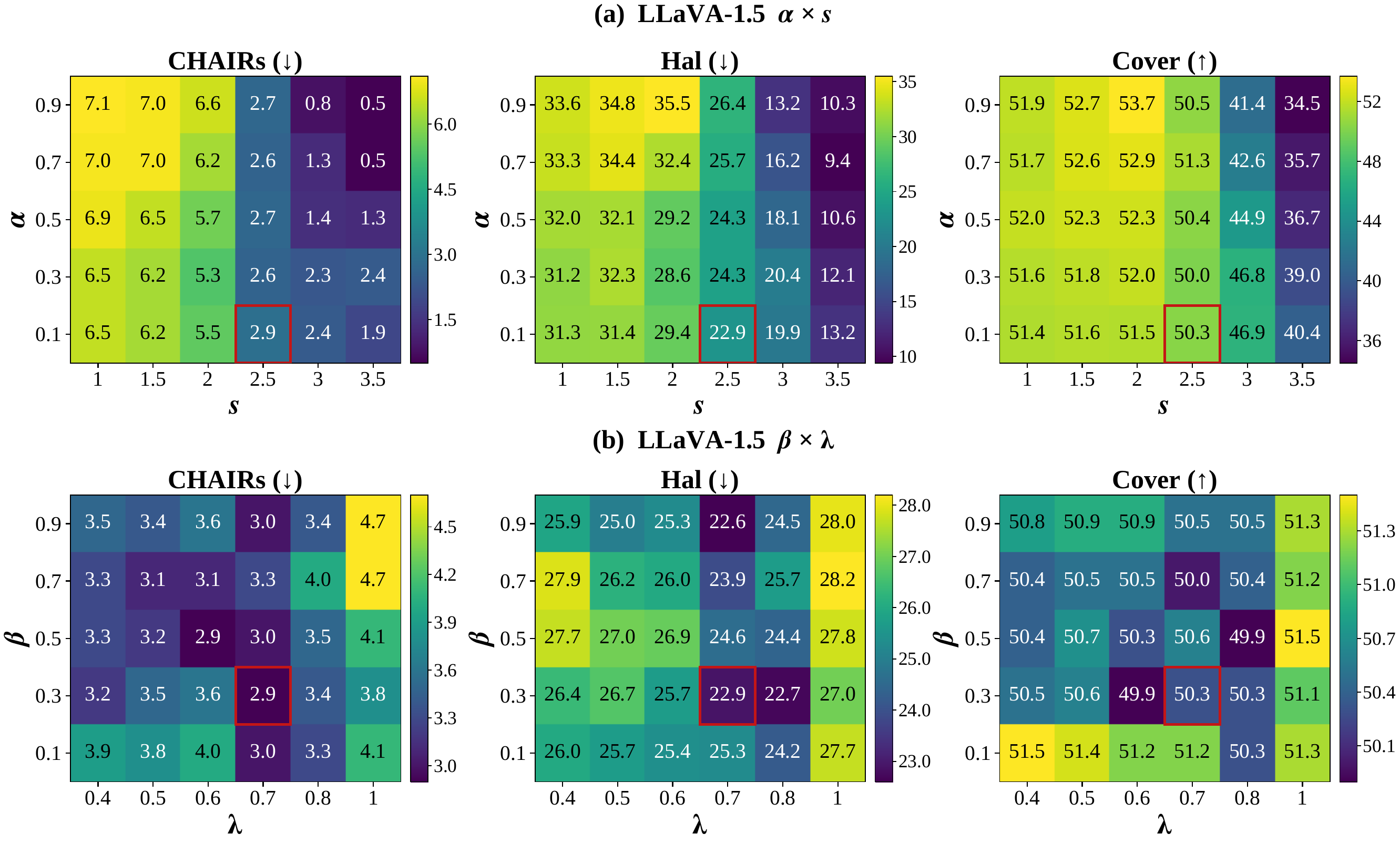}
    \par\vspace{-8pt} 

   \caption{\textbf{Hyperparameter sensitivity analysis on LLaVA-1.5.} We jointly examine $(\alpha,s)$ and $(\beta,\lambda)$. Stronger steering reduces hallucination but may compromise object coverage. A favorable balance occurs near $\alpha=0.1$, $s=2.5$, $\beta=0.3$, and $\lambda=0.7$.}
    \label{fig:hyperparameters}
    
\end{figure}

\noindent\textbf{Effects of Hyperparameters.} To evaluate the effects of hyperparameter settings on model performance, we examine two parameter combinations, $\alpha\times s$ and $\beta\times\lambda$, on LLaVA-1.5. As shown in Fig.~\ref{fig:hyperparameters}, increasing the steering strength \(s\) generally lowers hallucination metrics, but excessive intervention substantially reduces ground-truth object coverage. At larger \(s\), increasing \(\alpha\) usually further suppresses hallucination while exacerbating the decline in Cover, indicating that reference direction adjustment and injection magnitude must be coordinated. By contrast, \(\beta\) and \(\lambda\) have relatively minor effects on Cover, whereas hallucination metrics vary non-monotonically; neither relaxing the gate nor increasing the weight of the current vector yields consistent gains. Balancing hallucination suppression and information coverage, we use \(\alpha=0.1\), \(s=2.5\), \(\beta=0.3\), and \(\lambda=0.7\) as the main experimental configuration for LLaVA-1.5.

\begin{table}[t]
\centering

\caption{\textbf{Component ablations on AMBER with LLaVA-1.5 (unit: \%).} We evaluate the incremental integration of GVDS (dynamic direction construction), DGM (dimension-wise modulation), and DAAG (adaptive gating). The blue row denotes the complete DARES configuration.}
\label{tab:ablation}
\fontsize{6.5}{7.5}\selectfont
\renewcommand{\arraystretch}{0.85}
\setlength{\tabcolsep}{1pt}
\begin{tabularx}{\columnwidth}{@{}*{4}{c}*{5}{>{\centering\arraybackslash}X}@{}}
\toprule
\multirow{2}{*}[-0.5ex]{Baseline} & \multirow{2}{*}[-0.5ex]{GVDS} & \multirow{2}{*}[-0.5ex]{DGM} & \multirow{2}{*}[-0.5ex]{DAAG} & \multicolumn{5}{c}{AMBER} \\
\cmidrule(lr){5-9}
& & & & CHAIR\(\downarrow\) & Cover\(\uparrow\) & Hal\(\downarrow\) & Cog\(\downarrow\) & Hal.~Avg.\(\downarrow\) \\
\midrule
\(\checkmark\) & & & & 7.3 & 50.9 & 33.2 & 3.9 & 14.8 \\
\(\checkmark\) & \(\checkmark\) & & & 4.7 & 17.4 & 6.5 & 0.2 & 3.8 \\
\(\checkmark\) & \(\checkmark\) & \(\checkmark\) & & 2.2 & 28.2 & 4.6 & 0.2 & 2.3 \\
\rowcolor{blue!5}
\(\checkmark\) & \(\checkmark\) & \(\checkmark\) & \(\checkmark\) & 2.9 & 50.3 & 22.9 & 2.1 & 9.3 \\
\bottomrule
\end{tabularx}
\par\vspace{-8pt}
\end{table}

\begin{figure}[t]
    \centering
    \includegraphics[width=\columnwidth]{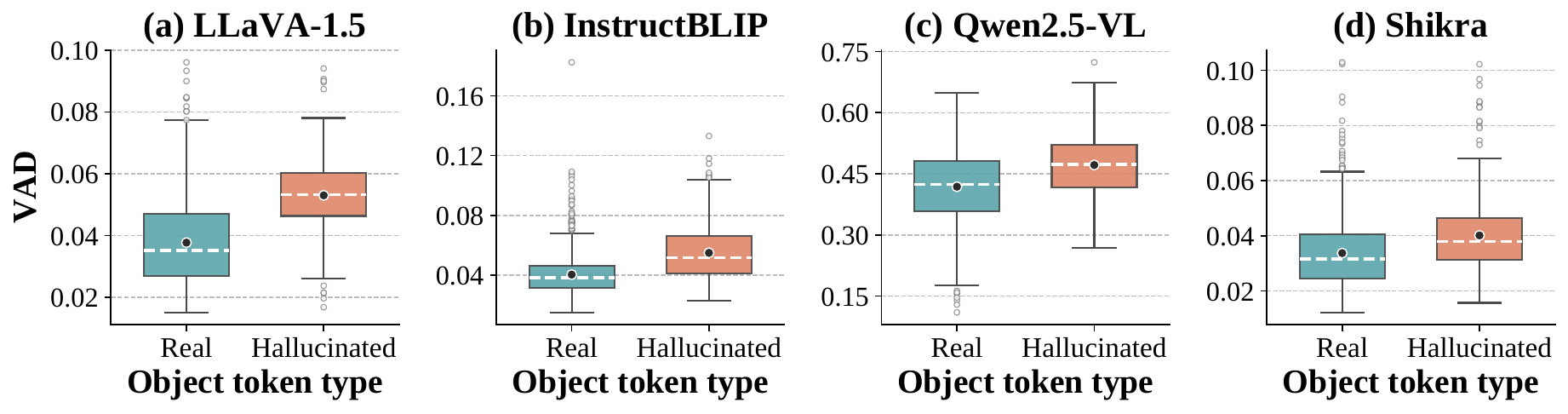}
    \par\vspace{-8pt} 

    \caption{\textbf{Shallow-layer VAD distributions across four LVLMs.} Hallucinated object tokens consistently exhibit higher central VAD values than grounded object tokens, while absolute score scales vary across backbones. These distributions support the use of shallow-layer VAD for hallucination risk monitoring.}
    \label{fig:shallow_vad}
    \par\vspace{-8pt}
\end{figure}

\noindent\textbf{Association Between Shallow-Layer VAD and Hallucination.} To further examine the association between shallow-layer VAD and object hallucination, as well as its cross-model applicability, Fig.~\ref{fig:shallow_vad} compares the L0--L2 VAD distributions of grounded and hallucinated object tokens in LLaVA-1.5, InstructBLIP, Qwen2.5-VL, and Shikra. Across all four backbones, the distributions of hallucinated object tokens exhibit a higher central tendency and shift toward larger VAD values, indicating greater attention deviation from the instruction--vision reference. This consistent positive association across models further supports shallow-layer VAD as a signal for monitoring hallucination risk. Marked differences in score scales across backbones also motivate adaptive thresholds for intervention timing.

\begin{figure}[t]
    \centering
    \includegraphics[width=\columnwidth]{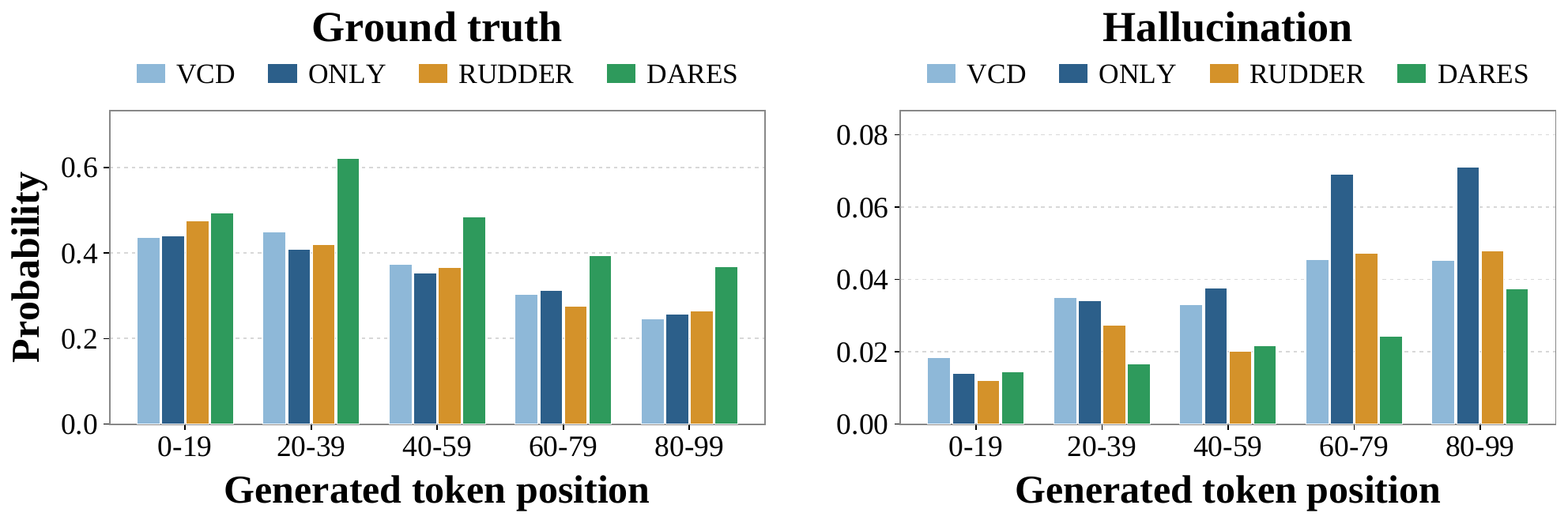}
    \par\vspace{-8pt} 

   \caption{\textbf{Object prediction probabilities across generated token positions.} Probabilities conditioned on noun generation are shown for grounded (left) and hallucinated (right) objects. Hallucination probabilities generally rise with generation for VCD and ONLY, while RUDDER still shows a late-stage increase. DARES maintains higher grounded-object probabilities and the lowest hallucination probabilities at positions 60--99.}
    \label{fig:long_term_decay}
    \par\vspace{-8pt}
\end{figure}

\noindent\textbf{Long-Term Decay.} Recent studies show that as autoregressive generation progresses, the influence of language priors increases while visual grounding weakens. RUDDER and FLB mitigate this issue by continuously reusing early visual guidance signals, demonstrating the advantage of steering methods in preserving visual evidence over long sequences. To investigate whether DARES retains and extends this advantage, we follow prior analyses and adopt a probability metric conditioned on noun generation to compare the prediction probabilities of grounded and hallucinated objects across decoding positions. As shown in Fig.~\ref{fig:long_term_decay}, the conditional probabilities of grounded objects for non-steering methods (VCD and ONLY) generally decline as generation progresses, whereas those for hallucinated objects increase. This indicates their failure to effectively suppress hallucination induced by long-term decay. By contrast, RUDDER effectively mitigates hallucination during early and middle generation stages, but hallucination probabilities still rise markedly at later stages. DARES not only maintains higher grounded-object probabilities across all position intervals but also achieves the lowest hallucination probabilities at later positions (60--99). These results demonstrate that DARES preserves the advantage of steering methods in sustaining visual grounding while strengthening late-stage hallucination suppression, confirming its effectiveness in mitigating the long-term decay of visual grounding.

\begin{figure}[t]
    \centering
    \includegraphics[width=\columnwidth]{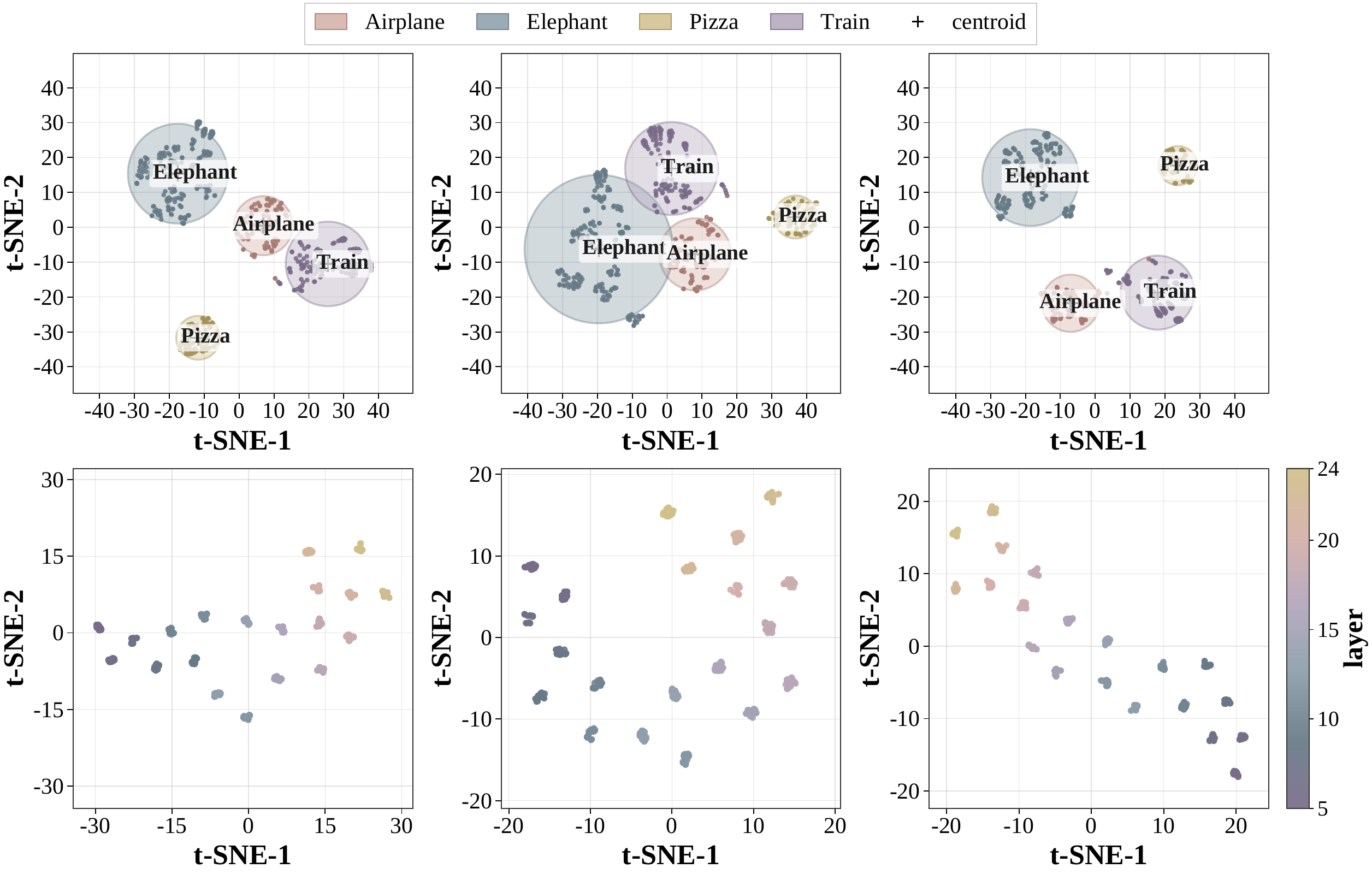}

    \caption{\textbf{t-SNE visualizations of DARES steering vectors.} Top: Vector distributions for four COCO object categories at L6, L14, and L22. Category clustering indicates that the steering vectors preserve visual semantic structure. Bottom: Vector distributions for different tokens across L5--L24. Inter-layer clustering qualitatively supports token-wise updated, layer-adaptive steering.}
    \label{fig:dynamic_tsne}
    \par\vspace{-8pt}
\end{figure}

\noindent\textbf{Dynamic and Fine-Grained Steering in DARES.} To examine how the proposed steering vectors adapt to the generation state, we first use t-SNE \cite{van2008visualizing} to visualize steering vectors for different tokens across layers L5--L24. As shown in Fig.~\ref{fig:dynamic_tsne}, vectors for different tokens within the same layer vary within local neighborhoods, whereas vectors from different layers form relatively distinct clusters, reflecting layer specificity. We then select layers L6, L14, and L22 and separately visualize steering vectors associated with four different COCO object categories. At all three selected layers, vectors from the same category tend to cluster together, while different categories exhibit distinguishable distributions, indicating that the steering vectors retain object category-related semantic structure rather than random activation noise. This category-level organization suggests that meaningful visual semantics are preserved across the representations of the selected layers. These results are consistent with steering directions being constructed per token during generation and adapted to each layer's representations, thereby providing qualitative support for the dynamic, fine-grained design of DARES. 

\section{Conclusion}
In this work, we revisit hallucination from the perspective of differences in visual attention distributions and introduce VAD, a novel attention-dynamics-aware metric. Our analysis reveals the association between visual discrepancy and hallucination, as well as its multi-granularity variations, which directly motivates DARES, a balanced, training-free framework for fine-grained dynamic visual residual steering. By combining adaptive gating, dynamic direction construction, and dimension-wise modulation, DARES enables highly selective, token-evolving, and layer-adaptive visual interventions. Extensive experiments demonstrate that our method achieves strong hallucination mitigation performance and cross-model adaptability.
{
    \small
    \bibliographystyle{ieeenat_fullname}
    \bibliography{main}
}
\clearpage
\section{Benchmark Details}
\label{sec:benchmark_details}

We evaluate object hallucinations on four benchmarks using greedy decoding with a maximum of 512 generated tokens. All metrics are reported as percentages except for Num, which is a count.

\noindent\textbf{AMBER.}

We use the generative split of AMBER, which contains 1,004 images covering 337 object types across 14 broad categories. The benchmark uses the prompt ``Describe this image.'' and evaluates each description against the annotations for that image. We report four benchmark metrics and a composite hallucination score:

\begin{itemize}
    \setlength{\itemsep}{0pt}
   
    \item \textbf{CHAIR} --- Percentage of extracted object mentions unsupported by the image annotations.
   
    \item \textbf{Cover} --- Percentage of annotated objects mentioned in the description.
    
    \item \textbf{Hal} --- Percentage of descriptions containing at least one hallucinated object.
    
    \item \textbf{Cog} --- Measures how often the model mentions objects annotated as plausible but absent from the image.
   
    \item \textbf{Hal.~Avg.} --- Arithmetic mean of the three hallucination metrics, $(\mathrm{CHAIR}+\mathrm{Hal}+\mathrm{Cog})/3$. Cover is reported separately.
\end{itemize}

\noindent\textbf{CHAIR.}

CHAIR evaluates object hallucinations in image captions against MSCOCO annotations. The evaluation protocol uses 500 images randomly sampled from the COCO 2014 validation set and the prompt ``Please describe the image in detail.'' We report:

\begin{itemize}
    \setlength{\itemsep}{0pt}

    \item $\boldsymbol{\mathrm{CHAIR}_{s}}$ --- Percentage of captions containing at least one hallucinated object.

    \item $\boldsymbol{\mathrm{CHAIR}_{i}}$ --- Percentage of all generated object mentions that are hallucinated.

    \item \textbf{Hal.~Avg.} --- Arithmetic mean of the two hallucination rates, $(\mathrm{CHAIR}_{s}+\mathrm{CHAIR}_{i})/2$.
\end{itemize}

\noindent\textbf{OPOPE.}

OPOPE adapts POPE's object presence evaluation to open-ended captioning. Instead of asking a yes/no question about each sampled object, it checks whether the generated caption mentions that object and compares the result with its annotated presence in the image. Each metric is averaged over the random, popular, and adversarial sampling settings:

\begin{itemize}
    \setlength{\itemsep}{0pt}

    \item \textbf{Accuracy (Acc)} --- Percentage of sampled objects whose presence or absence is correctly identified.

    \item \textbf{Precision (Pre)} --- Percentage of mentioned sampled objects that are present in the image.

    \item $\boldsymbol{F_{\beta=0.2}}$ --- Weighted harmonic mean of precision $P$ and recall $R$:
    \[
        F_{\beta}
        = \frac{(1+\beta^2)PR}{\beta^2 P+R},
        \qquad \beta=0.2.
    \]
    This score emphasizes precision because a faithful caption need not mention every object in the image.
\end{itemize}

\noindent\textbf{Object HalBench.}

Object HalBench evaluates detailed image descriptions on 300 COCO image--prompt pairs spanning eight prompt templates. A GPT-based extractor identifies object mentions in each generated caption, and these mentions are checked against COCO annotations. We report:

\begin{itemize}
    \setlength{\itemsep}{0pt}

    \item \textbf{C-s} --- Percentage of all captions containing at least one hallucinated COCO object.

    \item \textbf{C-SR} --- Uses the same numerator as C-s but restricts the denominator to captions mentioning at least one COCO object.

    \item \textbf{C-i} --- Percentage of extracted COCO object mentions that are hallucinated.

    \item \textbf{Num} --- Number of captions mentioning at least one COCO object, which serves as the denominator of C-SR.
\end{itemize}

\begin{figure*}[t]
    \centering
    \includegraphics[width=\textwidth]{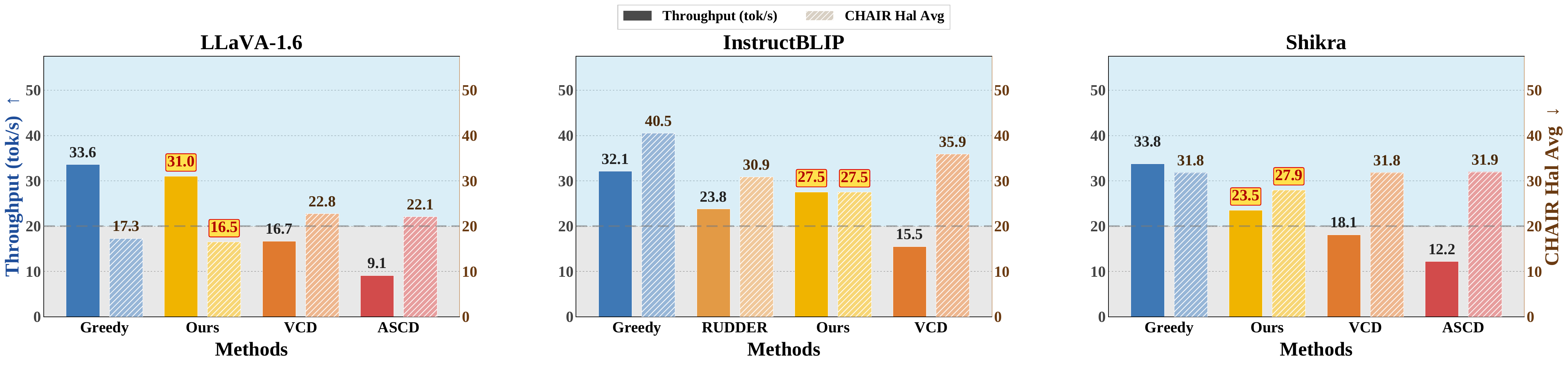}
    \caption{\textbf{Inference efficiency and hallucination on three additional
    backbones.} Solid bars show generation throughput (tokens/s; higher is
    better), and hatched bars show CHAIR Hal.\ Avg.\ (lower is better).
    On LLaVA-NeXT (LLaVA-1.6), InstructBLIP, and Shikra, DARES achieves the
    lowest hallucination score and the highest throughput among the mitigation
    methods compared.}
    \label{fig:efficiency_additional}
\end{figure*}

\section{Efficiency Comparisons}

We extend our efficiency analysis to LLaVA-NeXT, InstructBLIP, and Shikra, comparing generation throughput and Hal.\ Avg.\ on CHAIR.
As shown in Fig.~\ref{fig:efficiency_additional}, DARES achieves the lowest Hal.\ Avg.\ on all three backbones and the highest throughput among the mitigation methods compared.
Its throughput is $1.86\times$, $1.77\times$, and $1.30\times$ that of VCD on the three backbones, respectively.
On LLaVA-NeXT, DARES retains 92.3\% of greedy decoding throughput while reducing Hal.\ Avg.\ from 17.3 to 16.5.
On InstructBLIP, DARES also improves throughput by 15.5\% over RUDDER while achieving a lower Hal.\ Avg.
These results further support the trade-off DARES achieves between hallucination mitigation and inference efficiency across backbones.

\begin{table}[t]
    \centering
    \caption{Average word and token counts of captions generated by vanilla
    decoding and DARES on AMBER. DARES maintains a caption length comparable
    to that of the baseline.}
    \label{tab:amber_generation_length}
    \renewcommand{\arraystretch}{1.1}
    \begin{tabularx}{\columnwidth}{l *{2}{>{\centering\arraybackslash}X}}
        \toprule
        Method & Average words & Average tokens \\
        \midrule
        Vanilla & 75.64 & 96.96 \\
        \rowcolor[rgb]{0.9412,0.9569,1.0}
        DARES   & 74.45 & 95.42 \\
        \bottomrule
    \end{tabularx}
\end{table}

\section{Qualitative Analysis}

DARES selectively injects visual guidance into the residual stream during decoding. To assess whether hallucination reduction comes at the cost of substantially shorter captions, we compare the mean word and token counts of captions generated on AMBER with and without DARES. As shown in Table~\ref{tab:amber_generation_length}, DARES reduces the mean word count from 75.64 to 74.45 and the mean token count from 96.96 to 95.42, a decrease of approximately 1.6\% in both cases. Together with the lower hallucination scores on AMBER, these results indicate that DARES mitigates object hallucinations while preserving a caption length comparable to that of the base model.

\section{Additional t-SNE Visualizations of Steering Vectors}

\begin{figure*}[tp]
    \centering
    \includegraphics[
        width=\textwidth,
        height=0.78\textheight,
        keepaspectratio
    ]{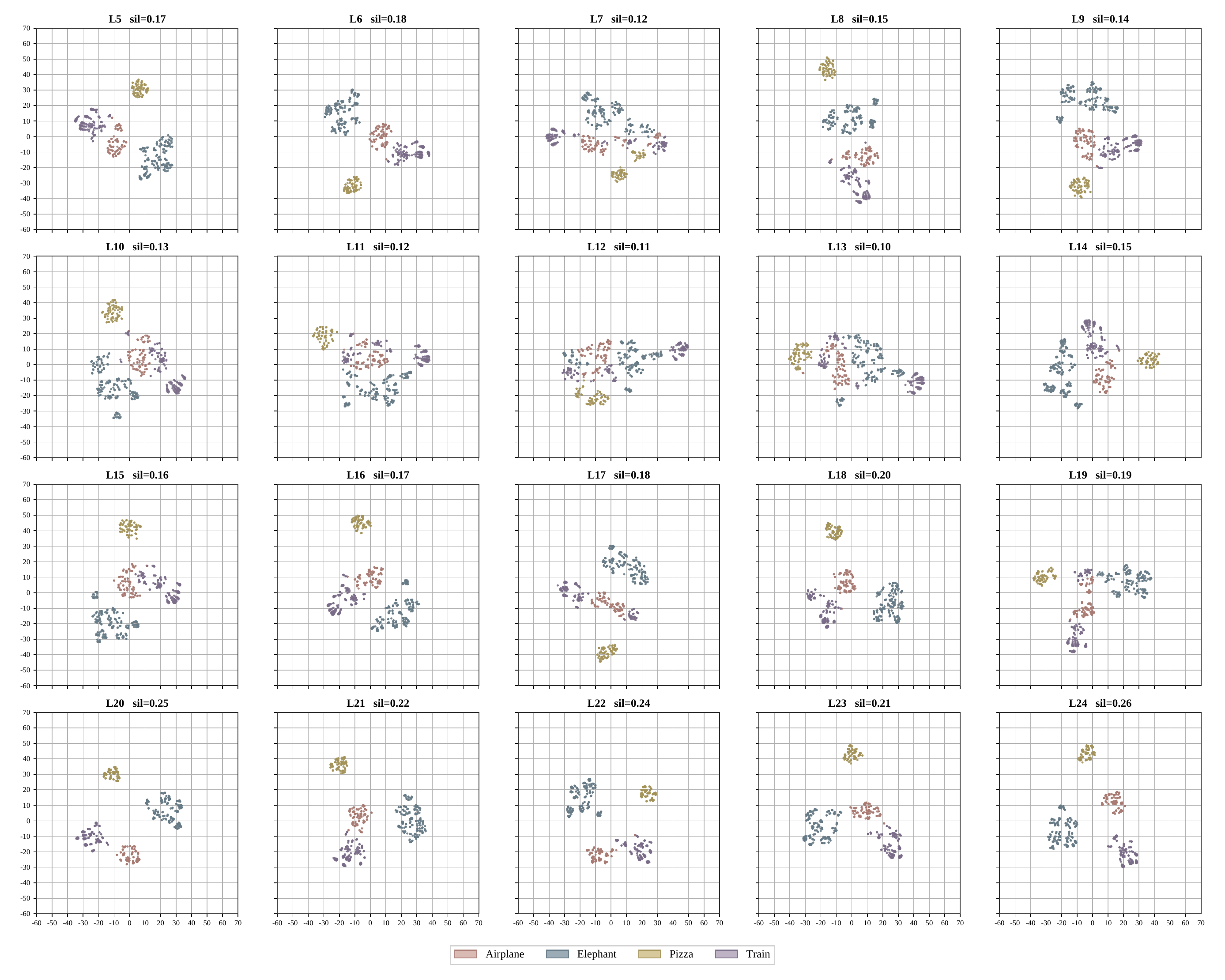}
    \caption{\textbf{Category structure of DARES steering vectors across
    injection layers.} Each panel shows a t-SNE projection for one layer
    from L5 to L24. Colors denote four COCO object categories: airplane,
    elephant, pizza, and train. Crosses mark category centroids, and panel
    titles report silhouette scores.}
    \label{fig:tsne_by_category}
\end{figure*}

\begin{figure*}[tp]
    \centering
    \includegraphics[width=\textwidth]{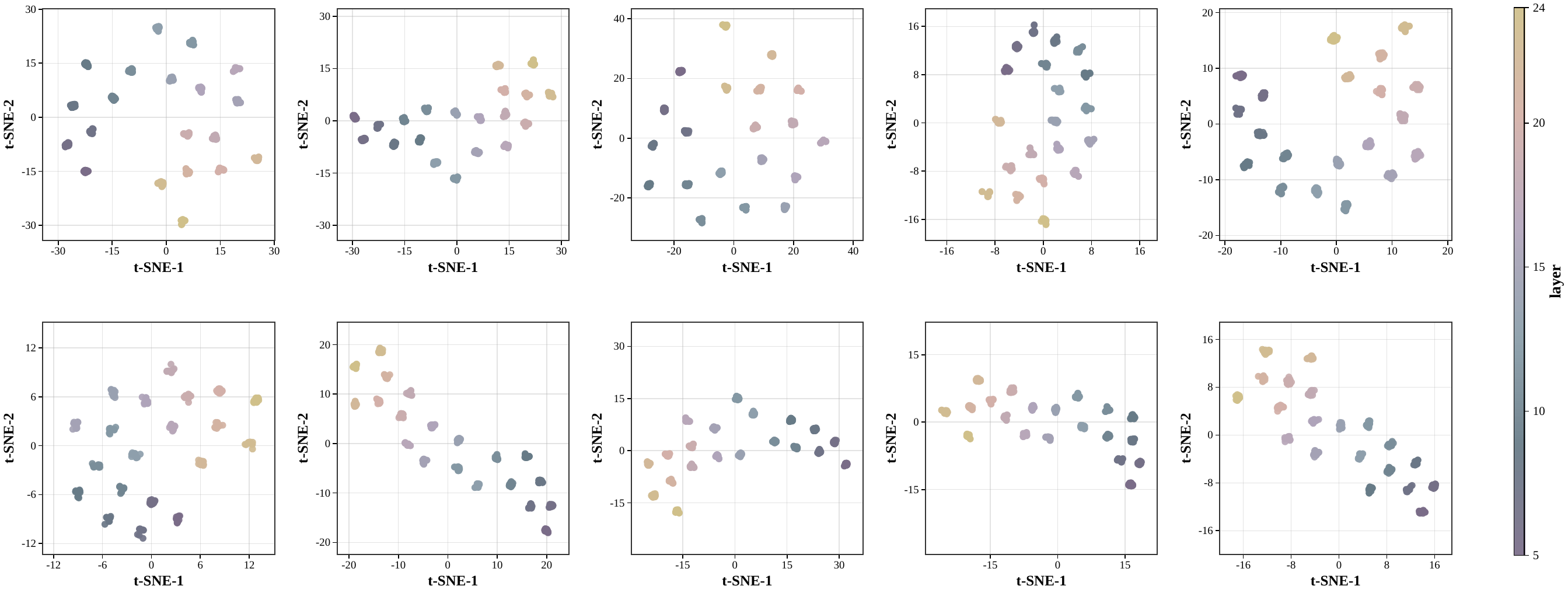}
    \caption{\textbf{Layer structure of DARES steering vectors across
    generated tokens.} Each point represents a token's steering vector
    at one injection layer (L5--L24). When colored by layer, vectors
    from the same layer tend to cluster together across tokens.}
    \label{fig:tsne_by_layer}
\end{figure*}

We extend the t-SNE analysis of L6, L14, and L22 to all 20 injection layers (L5--L24). As shown in Fig.~\ref{fig:tsne_by_category}, steering vectors for four COCO object categories (airplane, elephant, pizza, and train) tend to cluster by category. At L20--L24, the four categories form distinct clusters, with silhouette scores ranging from 0.21 to 0.26. Figure~\ref{fig:tsne_by_layer} shows the same-layer structure across tokens: vectors from the same layer tend to group together, forming clusters that are distinguishable by layer. Together, these patterns suggest that DARES steering vectors retain category-related visual semantics and reflect the representations of individual layers, providing qualitative support for layer-adaptive steering.

\section{Parameter Sensitivity}

\begin{figure*}[tp]
    \centering
    \includegraphics[width=\textwidth]
        {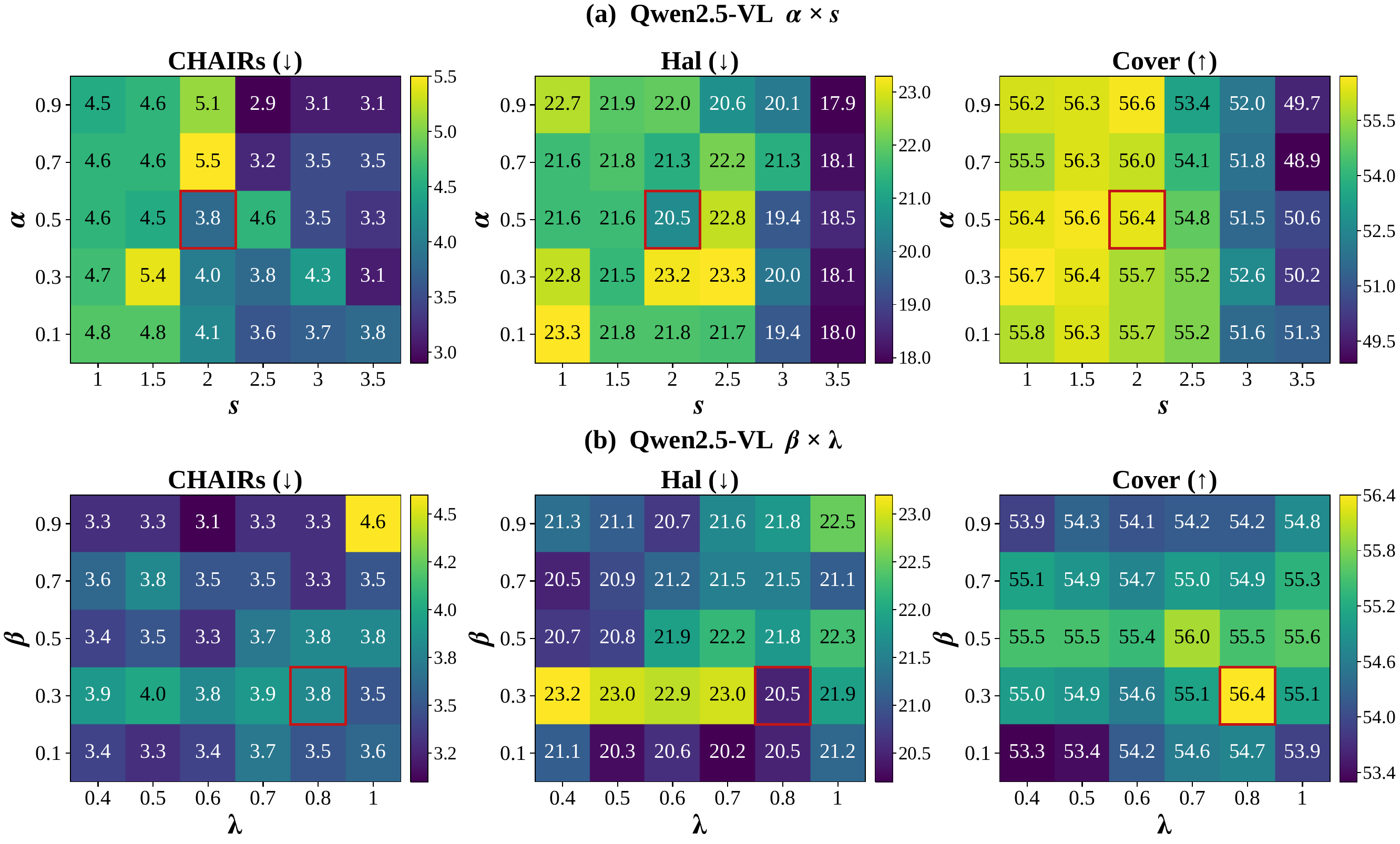}
    \caption{\textbf{Hyperparameter sensitivity on Qwen2.5-VL (AMBER).}
    (a) Joint effects of the reference coefficient $\alpha$ and steering
    strength $s$. (b) Joint effects of the EMA update coefficient $\beta$
    and gating threshold coefficient $\lambda$. Red boxes mark the main
    configuration: $(\alpha,s)=(0.5,2.0)$ and $(\beta,\lambda)=(0.3,0.8)$.}
    \label{fig:hp_qwen25}
\end{figure*}

\begin{figure*}[tp]
    \centering
    \includegraphics[width=\textwidth]
        {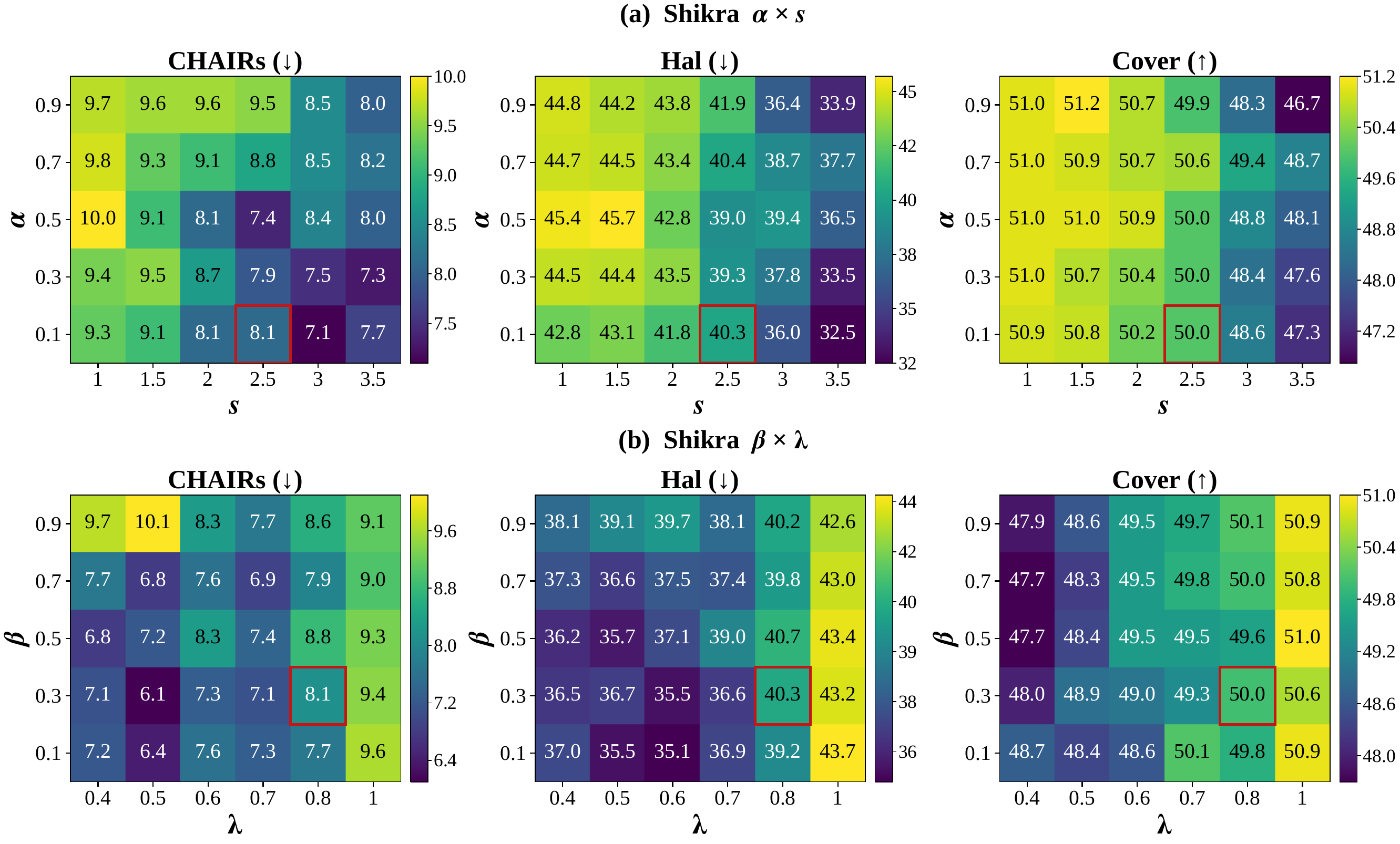}
    \caption{\textbf{Hyperparameter sensitivity on Shikra (AMBER).}
    (a) Joint effects of $\alpha$ and $s$.
    (b) Joint effects of $\beta$ and $\lambda$.
    Red boxes mark the main configuration: $(\alpha,s)=(0.1,2.5)$
    and $(\beta,\lambda)=(0.3,0.8)$.}
    \label{fig:hp_shikra}
\end{figure*}

\begin{figure*}[tp]
    \centering
    \includegraphics[width=\textwidth]
        {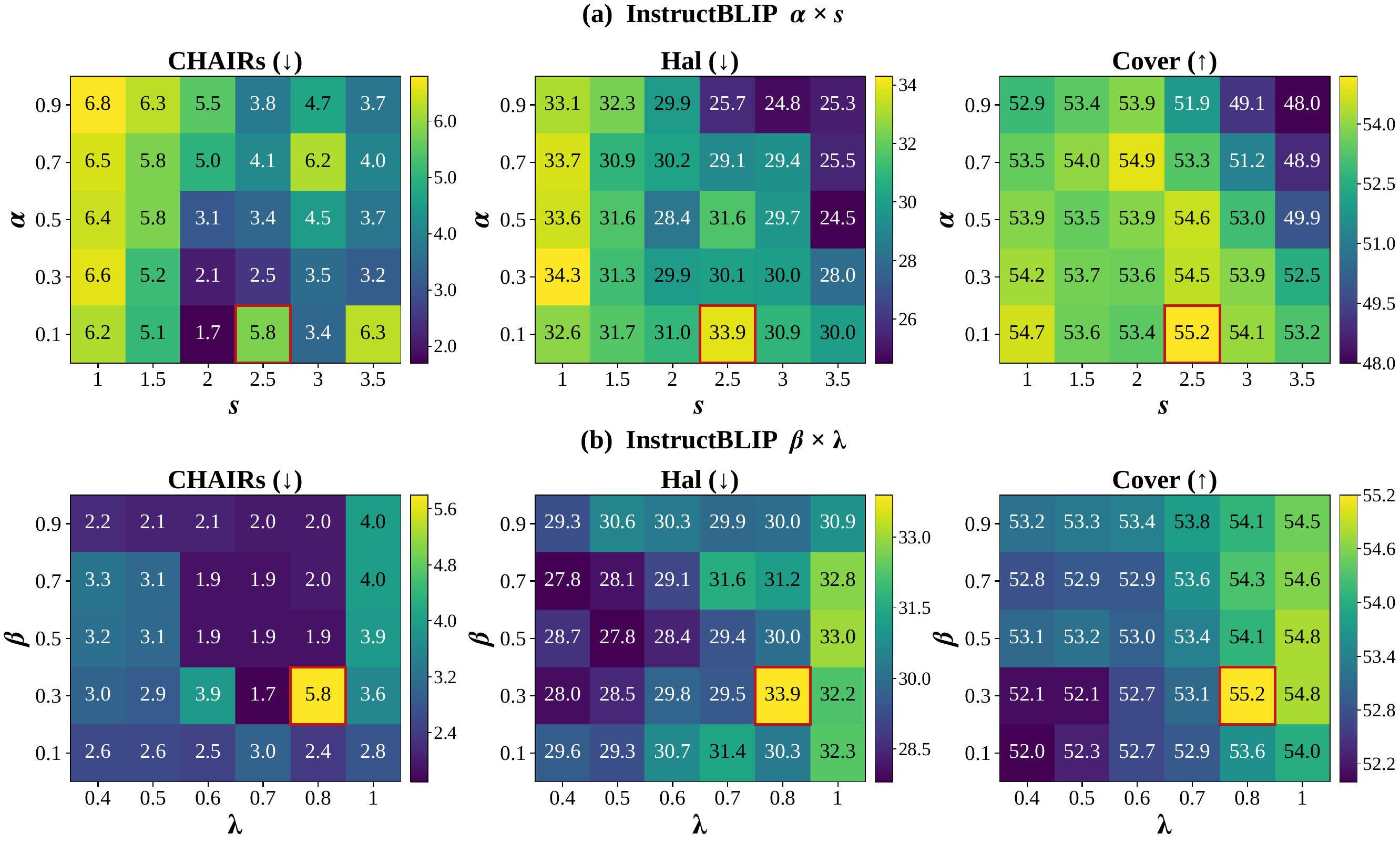}
    \caption{\textbf{Hyperparameter sensitivity on InstructBLIP (AMBER).}
    (a) Joint effects of $\alpha$ and $s$.
    (b) Joint effects of $\beta$ and $\lambda$.
    Red boxes mark the main configuration: $(\alpha,s)=(0.1,2.5)$
    and $(\beta,\lambda)=(0.3,0.8)$.}
    \label{fig:hp_instructblip}
\end{figure*}

\noindent \textbf{Hyperparameter Sensitivity Analysis.}
We extend the AMBER sensitivity analysis to Qwen2.5-VL, Shikra, and
InstructBLIP (Figs.~\ref{fig:hp_qwen25}, \ref{fig:hp_shikra},
and \ref{fig:hp_instructblip}). With the reference coefficient $\alpha$
fixed at each backbone's main setting, increasing the steering strength
$s$ from its default value to 3.5 lowers Hal on all three backbones,
but also reduces object coverage. For example, on Qwen2.5-VL
($\alpha=0.5$), increasing $s$ from 2.0 to 3.5 lowers Hal from 20.5
to 18.5 and Cover from 56.4 to 50.6. The reference coefficient also
affects this trade-off: on InstructBLIP at $s=3.5$, raising $\alpha$
from 0.1 to 0.9 lowers Hal from 30.0 to 25.3 and Cover from 53.2 to 48.0.

The joint sweep of the EMA update coefficient $\beta$ and the gating
threshold coefficient $\lambda$ reveals a backbone-dependent response.
At $\beta=0.3$, increasing $\lambda$ from 0.6 to 0.8 improves Cover on
all three backbones, whereas Hal decreases on Qwen2.5-VL but increases
on Shikra and InstructBLIP. Balancing hallucination reduction and object
coverage, we use $(\beta,\lambda)=(0.3,0.8)$ for all three backbones
and set $(\alpha,s)$ to $(0.5,2.0)$ for Qwen2.5-VL and $(0.1,2.5)$
for Shikra and InstructBLIP.

\noindent \textbf{Steering Layer Range.}
We evaluate DARES on AMBER with different ranges of residual injection layers.
As shown in Table~\ref{tab:steering_layer_range}, when steering starts at L5, extending the range from L5--12 to L5--24 reduces Hal.\ Avg.\ from 13.7 to 9.3 on LLaVA-1.5 and from 9.1 to 8.4 on Qwen2.5-VL.
Restricting steering to L0--4 or L25--27 also yields a higher Hal.\ Avg.\ on both backbones.
Among the ranges shown, L5--24 achieves the lowest Hal.\ Avg.\ on both models, with object coverage of 50.3\% and 56.4\%, respectively. This range coincides with the layers examined in our VAD analysis (F2), and we adopt it as the default injection range.

\begin{table*}[t]
  \centering
  \caption{\textbf{Effect of the steering layer range on AMBER (\%).}
  Hal.\ Avg.\ is the arithmetic mean of CHAIR, Hal, and Cog.
  Higher Cover and lower values on all other metrics are better. Bold denotes the best result in each column, including ties.
  The shaded row indicates the default range used in the main experiments.}
  \label{tab:steering_layer_range}
  \small
  \setlength{\tabcolsep}{3pt}
  \renewcommand{\arraystretch}{1.13}
  \begin{tabularx}{\textwidth}{l*{10}{>{\centering\arraybackslash}X}}
    \toprule
    & \multicolumn{5}{c}{\textbf{LLaVA-1.5}}
    & \multicolumn{5}{c}{\textbf{Qwen2.5-VL}} \\
    \cmidrule(lr){2-6}\cmidrule(lr){7-11}
    \textbf{Layers}
    & CHAIR & Cover & Hal & Cog & Hal.\ Avg.
    & CHAIR & Cover & Hal & Cog & Hal.\ Avg. \\
    \midrule
    L0--4  & 7.4 & 51.6 & 36.5 & 4.1 & 16.0 & 5.8 & 55.7 & 24.5 & 1.4 & 10.6 \\
    L5--12 & 6.4 & 52.0 & 31.5 & 3.3 & 13.7 & 4.7 & 56.0 & 21.7 & \textbf{1.0} & 9.1 \\
    L5--16 & 3.9 & 50.5 & 25.9 & 2.5 & 10.8 & 4.7 & 54.8 & 22.5 & 1.2 & 9.5 \\
    L8--24 & 4.4 & 51.6 & 27.1 & 2.7 & 11.4 & 4.3 & 56.0 & 23.3 & 1.1 & 9.6 \\
    L16--24 & 6.6 & \textbf{52.4} & 32.0 & 3.5 & 14.0
            & 4.7 & \textbf{57.1} & 23.6 & 1.2 & 9.8 \\
    L25--27 & 7.4 & 51.2 & 34.9 & 3.9 & 15.4 & 4.9 & 56.8 & 23.2 & 1.3 & 9.8 \\
    \rowcolor[rgb]{0.9412,0.9569,1.0}
    \textbf{L5--24} & \textbf{2.9} & 50.3 & \textbf{22.9} & \textbf{2.1} & \textbf{9.3}
                    & \textbf{3.8} & 56.4 & \textbf{20.5} & \textbf{1.0} & \textbf{8.4} \\
    \bottomrule
  \end{tabularx}
\end{table*}

\section{Head Fusion Weights and Dimension-wise Steering}
\label{sec:head_dimension_steering}

\begin{figure*}[p]
    \centering
    \includegraphics[
        width=\textwidth,
        height=0.86\textheight,
        keepaspectratio
    ]{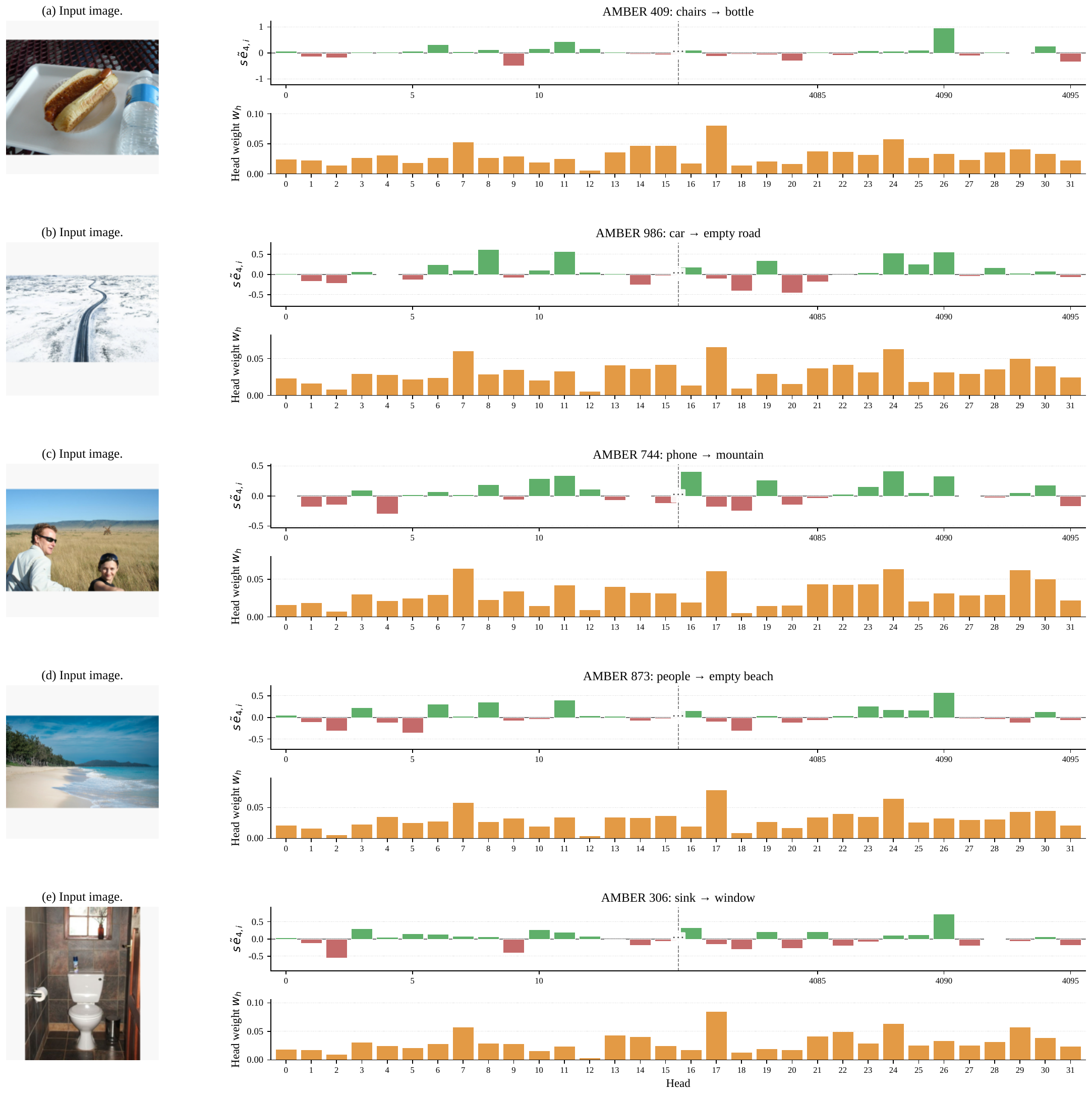}
    \caption{\textbf{Head fusion weights and dimension-wise steering on AMBER.}
    Each row shows an input image and the corresponding steering statistics.
    Upper plots show signed steering components for selected hidden dimensions
    (green: positive; red: negative); the axis break omits intervening dimensions.
    Lower plots show GVDS fusion weights across all 32 attention heads.
    Several heads consistently receive relatively high weights, while their
    relative contributions and the magnitudes of individual steering components
    vary across examples.}
    \label{fig:head_dimension_steering}
\end{figure*}

To examine how DARES steers generation across attention heads and hidden
dimensions, we visualize head fusion weights and dimension-wise steering
components for five AMBER examples. As shown in
Fig.~\ref{fig:head_dimension_steering}, fusion weights are distributed
unevenly across heads, and their relative magnitudes vary across examples.
Steering components likewise differ in sign and magnitude across hidden
dimensions, with distinct patterns across examples. These observations
illustrate the complementary roles of GVDS and DGM: GVDS weights each
head's contribution to the steering direction, while DGM modulates the
strength of the update along each hidden dimension.

\section{The DARES Algorithm}
\label{sec:supp_dares_algorithm}

Algorithm~\ref{alg:dares} illustrates DARES inference. Prefill builds the visual references once. At each decoding step, shallow layers determine whether to intervene; when the gate opens, each injection layer constructs and applies its own steering direction in the same forward pass.

\noindent\textbf{Notation.}
Let $\mathcal G$ and $\mathcal I$ denote the shallow gating layers and the subsequent injection layers, respectively.
$\mathcal N$ denotes $\ell_2$ normalization and $\mathcal M$ denotes per-map min--max normalization.
We use $D$ for visual attention discrepancy and $R_{\ell h}$ for the visual readout of head $h$ in layer $\ell$:
\[
D(\mathbf p,\mathbf q)
=1-\langle\mathcal N(\mathcal M(\mathbf p)),\mathcal N(\mathcal M(\mathbf q))\rangle,
\]
\[
R_{\ell h}(\mathbf p)=\mathbf W_O^{(\ell,h)}
  \sum_{j\in\mathcal J_v}p_j\mathbf v_j^{(\ell,h)}.
\]
Here $\mathcal J_v$ indexes visual tokens. During prefill, averaging IVA over heads gives the reference $\mathbf c_k$ for each gating layer $k$. Averaging per-head IVA over gating layers gives $\mathbf p_{\rm ins}^{(h)}$, whose head mean is the shared reference $\mathbf c$.
During decoding, $\mathbf A_t^{(k,h)}$ denotes the current token's visual attention map; $\mathbf p_t^{(h)}$ is its mean over gating layers. The visual readout uses raw attention weights. The backbone polarity $\sigma_b$ and coefficients $\lambda,\alpha,\beta,s$ are fixed during inference.

\ifdefined\SetAlgoLined
  \newenvironment{DaresWide}{\begin{algorithm*}[t]}{\end{algorithm*}}
\else
  \floatstyle{ruled}\newfloat{daresfloat}{tp}{ldares}\floatname{daresfloat}{Algorithm}
  \newenvironment{DaresWide}{\begin{daresfloat*}[t]}{\end{daresfloat*}}
\fi
\begin{DaresWide}
\caption{Dynamic Discrepancy-Aware Residual Steering (DARES)}
\label{alg:dares}
\noindent\begin{minipage}{\linewidth}
\normalsize
\setlength{\parindent}{0pt}
\setlength{\parskip}{0pt}
\setlength{\abovedisplayskip}{3pt}
\setlength{\belowdisplayskip}{3pt}
\setlength{\abovedisplayshortskip}{3pt}
\setlength{\belowdisplayshortskip}{3pt}
\setlength{\jot}{2pt}
\newcommand{\DaresStep}[1]{\par\addvspace{0.4\baselineskip}\noindent\textbf{#1}\par\nobreak}
\textbf{Input:} Image, instruction, frozen LVLM, and layer sets $\mathcal G,\mathcal I$.\par
\textbf{Output:} Generated token sequence $Y$.\par

\DaresStep{1. Cache visual references during prefill.}
Cache the KV states and IVA references. For each injection layer $\ell$, store its fixed reference direction:
\[
\mathbf r_{\rm ins}^{(\ell)}
\gets \mathcal N\!\left(\operatorname{avg}_{h} R_{\ell h}(\mathbf p_{\rm ins}^{(h)})\right).
\]
Set $Y=(y_1)$ using the greedy prefill prediction. Initialize empty history and set each smoothing state $\mathbf m_\ell$ to zero.\par

\par\addvspace{0.55\baselineskip}
\textit{Repeat Steps 2--5 for each decoding step until EOS or the length limit.}\par
\DaresStep{2. Determine whether to intervene (DAAG).}
Compute the shallow-layer discrepancy and compare it with a threshold based on the historical mean $\mu_t=\operatorname{avg}_{i<t}\overline D_i$:
\[
\begin{gathered}
\overline D_t \gets
\operatorname{avg}_{k\in\mathcal G}
D\!\left(\operatorname{avg}_h\mathbf A_t^{(k,h)},\mathbf c_k\right),\\
z_t \gets \mathbf 1[\overline D_t>\lambda\mu_t].
\end{gathered}
\]
With no history, leave the gate closed. If $z_t=0$, continue standard decoding and go to Step~5.\par

\DaresStep{3. Construct the steering direction (GVDS).}
For each head, compute its historical VAD $\delta_h=D(\operatorname{avg}_{i<t}\mathbf p_i^{(h)},\mathbf c)$.
Convert these scores to fusion weights:
\[
w_h=\frac{q_h}{\sum_{h'}q_{h'}+\varepsilon},\qquad
q_h=
\begin{cases}
\delta_h, & \sigma_b=-1,\\
\max(1-\delta_h,0), & \sigma_b=+1 .
\end{cases}
\]
At each $\ell\in\mathcal I$, fuse the current visual readouts and adjust the result using the fixed reference:
\[
\begin{gathered}
\mathbf r_t^{(\ell)} \gets \mathcal N\!\left(
   \sum_h w_h\,\mathcal N(R_{\ell h}(\mathbf p_t^{(h)}))\right),\\
\mathbf d_t^{(\ell)} \gets \mathcal N\!\left(
   \mathbf r_t^{(\ell)}+\sigma_b\alpha\,\mathbf r_{\rm ins}^{(\ell)}\right).
\end{gathered}
\]

\DaresStep{4. Modulate and inject the residual (DGM).}
Before advancing to the next layer, modulate and smooth the direction, then scale it to the current attention update:
\[
\begin{gathered}
\mathbf g_t^{(\ell)} \gets \tfrac12\!\left[\mathbf 1+
   \tanh\!\left(d_{\rm model}\,\mathbf d_t^{(\ell)}\odot\mathcal N(\mathbf a_{\ell,t})\right)\right],\\
\mathbf m_\ell \gets \beta\bigl(\mathbf g_t^{(\ell)}\odot\mathbf d_t^{(\ell)}\bigr)
   +(1-\beta)\mathbf m_\ell,\\
\mathbf a_{\ell,t} \gets \mathbf a_{\ell,t}
   +s\|\mathbf a_{\ell,t}\|_2\,\mathcal N(\mathbf m_\ell).
\end{gathered}
\]

\DaresStep{5. Update history and generate the next token.}
Update the running means with $\overline D_t$ and $\{\mathbf p_t^{(h)}\}_h$.
Append the greedy prediction $y_{t+1}$ to $Y$.\par
\end{minipage}
\end{DaresWide}

\section{Case Study}

We compare captions generated with and without DARES to examine how steering affects which objects the model describes.

\begin{figure*}[p]
    \centering
    \includegraphics[width=\textwidth,height=\textheight,keepaspectratio]
        {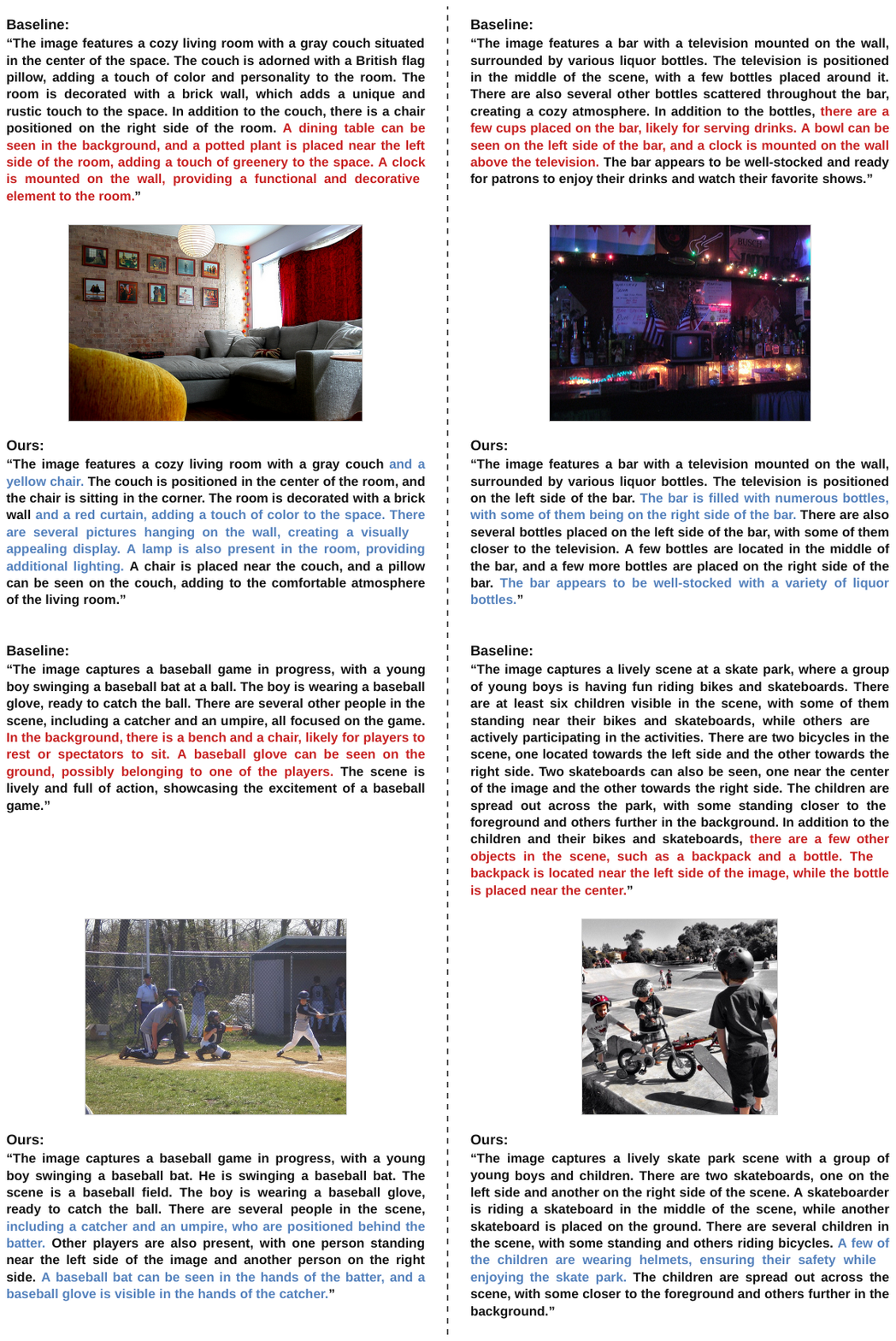}
    \par\vspace{-10pt}
    \caption{\textbf{Case studies on CHAIR.}
    Red: hallucinations from vanilla LLaVA-1.5. Blue: grounded details from DARES.}
    \label{fig:case_chair}
\end{figure*}

As shown in Fig.~\ref{fig:case_chair}, the baseline places a dining table in the background of the living room, a potted plant on the left, and a clock on the wall. DARES omits these unsupported objects and instead describes the red curtain and the framed pictures on the brick wall. Similarly, in the skate park scene, DARES mentions the children's helmets without introducing the backpack or bottle described by the baseline.

\begin{figure*}[p]
    \centering
    \includegraphics[width=\textwidth,height=\textheight,keepaspectratio]
        {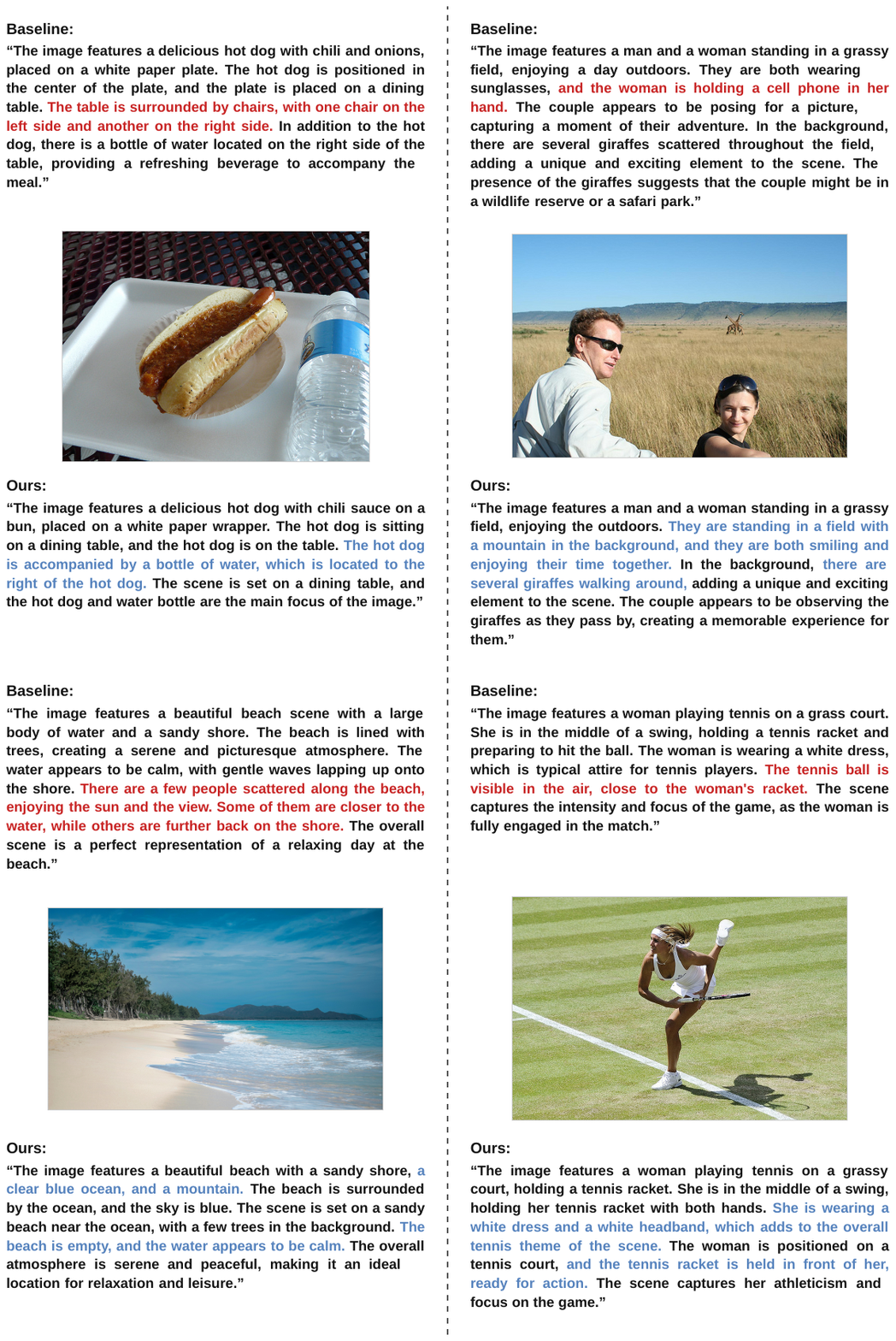}
    \par\vspace{-10pt}
    \caption{\textbf{Case studies on AMBER.}
    Red: hallucinations from vanilla LLaVA-1.5. Blue: grounded details from DARES.}
    \label{fig:case_amber}
\end{figure*}

The AMBER examples in Fig.~\ref{fig:case_amber} show a similar pattern. Where the baseline mentions a car, DARES describes an empty road through the snow; in another case, it identifies the bottle beside a hot dog rather than reporting chairs. Together, these examples illustrate how DARES can suppress plausible but unsupported object mentions while preserving details grounded in the image.

\end{document}